\pdfoutput=1

\documentclass[11pt]{article}

\usepackage{acl}

\usepackage{times}
\usepackage{amsthm}
\usepackage{latexsym}
\usepackage{graphicx}
\usepackage{color}
\usepackage{url}
\usepackage{amssymb}
\usepackage{amsmath}
\usepackage{multirow}
\usepackage{booktabs}
\usepackage{algorithm}
\usepackage{algpseudocode}
\usepackage{wrapfig}
\newcommand{\our} {\textsc{Fair}}
\newcommand{\snuba} {\textsc{Snuba}}
\newcommand{\spear} {\textsc{Spear}}
\newcommand{\astra} {\textsc{Astra}}
\newcommand{\grasp} {\textsc{Grasp}}

\usepackage[T1]{fontenc}

\usepackage[utf8]{inputenc}

\usepackage{microtype}

\usepackage{inconsolata}

\title{\our: Filtering of Automatically Induced Rules}

\author{Divya Jyoti Bajpai\textsuperscript{1}, Ayush Maheshwari\textsuperscript{2}\thanks{~~Outcome of research while pursuing PhD at IIT Bombay.}~, Manjesh Kumar Hanawal\textsuperscript{1}, \\ \textbf{Ganesh Ramakrishnan\textsuperscript{1}} \\
  \textsuperscript{1} Indian Institute of Technology Bombay, India ~~~  \textsuperscript{2} Vizzhy Inc, Bengaluru, India\\
\texttt{\{divyajyoti.bajpai, mhanawal, ganramkr\}@iitb.ac.in}\\
\texttt{ayush.maheshwari@vizzhy.com}
  }

\begin{document}
%
%
%
%
\maketitle              
\begin{abstract}
The availability of large annotated data can be a critical bottleneck in training machine learning algorithms successfully, especially when applied to diverse domains. 
Weak supervision offers a promising alternative by accelerating the creation of labeled training data using domain-specific rules. However, it requires users to write a diverse set of high-quality rules to assign labels to the unlabeled data. Automatic Rule Induction (ARI) approaches circumvent this problem by automatically creating rules from features on a small labeled set and filtering a final set of rules from them. In the ARI approach, the crucial step is to filter out a set of a high-quality useful subset of rules from the large set of automatically created rules. In this paper, we propose an algorithm \our{} (\textbf{F}iltering of \textbf{A}utomatically \textbf{I}nduced \textbf{R}ules) to filter rules from a large number of automatically induced rules using submodular objective functions that account for the collective precision, coverage, and conflicts of the rule set. We experiment with three ARI approaches and five text classification datasets to validate the superior performance of our algorithm with respect to several semi-supervised label aggregation approaches. Further, we show that \our{} achieves statistically significant results in comparison to existing rule-filtering approaches. The source code is available at \url{https://github.com/ayushbits/FAIR-LF-Induction}. 
\end{abstract}

\section{Introduction}
Machine learning applications rely on large amounts of labeled training data to obtain state-of-the-art performance on downstream tasks such as text classification, machine translation, image captioning, \textit{etc.}
However, it is expensive to obtain high-quality labeled training data. Therefore, several methods such as crowdsourcing \cite{brabham2013crowdsourcing}, self-supervision \cite{asano2019critical}, distant supervision \cite{mintz2009distant} and semi-supervision \cite{van2020survey} techniques have been proposed to reduce the human annotation efforts. 
Another popular technique, \textit{viz}, weak supervision, aims to quickly create labeled data by leveraging \textit{expert} defined rules. These rules are generic patterns developed by assessing a few exemplars from the corpus. Typically, users encode supervision as rules in the form of labeling functions (LFs), where each rule assigns a noisy label to an instance. However, these different rules can assign different labels to an instance. Weak supervision approaches aggregate and resolves these conflicting rules to assign a weak label to an instance. 

\begin{figure*}[!h]
    \centering
    \includegraphics[width=0.8\linewidth]{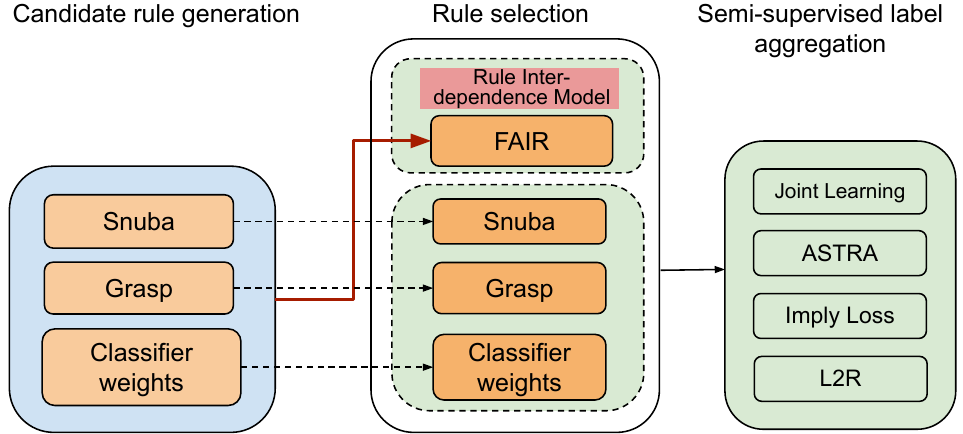}
     \caption{
    The flow of our approach. We first generate rules in the candidate rule generation block and then filter them using different respective approaches (such as with \snuba~\cite{snuba}, \grasp~\cite{shnarch2017Grasp}  and Classifier weights) as also with \our{}. The final committed rule set is passed on to the semi-supervised label aggregation approaches for the final performance on the downstream task.}
    \label{fig:flow1}
\end{figure*}

Although weak supervision methods reduce the data annotation effort, they still require human experts to frame and encode rules. Automatic rule induction (ARI) approaches circumvent this problem by automatically inducing rules from the data. ARI methods use a small labeled set to extract rules either by using decision tree approaches~\cite{snuba} or weights of a classifier~\cite{pryzant2022automatic}. Other ARI approaches such as \grasp~\cite{shnarch2017Grasp} extract rich linguistic patterns from a given set of positive and negative examples.
These approaches initially find and filter a list of patterns to find the top-\textit{k} patterns. These patterns are transformed into rules that yield noisy labels.
The rules are then fed to the unsupervised \cite{bach2019snorkel, oishik} or semi-supervised aggregation approaches  to\cite{maheshwari2022learning, astra} aggregate noisy labels.


Current ARI approaches select a final set of useful rules without considering explicit interdependencies between the rules. Classifier weights \cite{pryzant2022automatic} and M-\grasp\ \cite{shnarch2017Grasp} greedily select top-$k$ patterns having the highest weights assigned by the classifier. In this approach, any interdependence among LFs is not captured as LFs are based on the top-ranked features and not on their labeling properties. \snuba~\cite{varma2018snuba} chooses a rule in every iteration that maximizes the weighted sum of the F1 score on the labeled set and Jacard score. Then \snuba\ reduces the labeled set size by removing instances labeled by added rule. 
Since LFs are generated only on the partially labeled set, the dependency of the rules set is not explicitly captured. Further, \snuba\ is computationally involved since new LFs need to be generated iteratively depending upon the verifier's feedback. Also, this can lead to similar LFs getting added to the committed set, thereby causing instability.

As an illustration, for the question classification dataset which classifies a given question into five different classes, \snuba\ selects the following rules: \textit{how}, \textit{how many}, and \textit{many} for the class \textit{Numeric} while \our{}{ (\textbf{F}iltering of \textbf{A}utomatically \textbf{I}nduced \textbf{R}ules)} (Section~\ref{sec:rule-methods}) selects only the \textit{how} rule in conjunction with other independent rules\footnote{ A class label is associated with a rule, \emph{for eg}, if \textit{how} appears in the text, the rule assigns a weak label as \textit{numeric} for an instance.}. Clearly, derived rules for \textit{how}, belonging to the same class, are dependent on the parent rule. Due to its iterative procedure of working with new induced rules from the uncovered labeled set, \snuba\ selects LFs showing high overlap amongst themselves. Our algorithm captures explicit interdependence among LFs, thus selecting diverse and representative rules.



In this work, we propose a method, \our\, to select a subset of LFs ({\it committed set}) from a fixed set of automatically induced LFs ({\it candidate set}). Firstly, we consider a natural objective function defined as a weighted sum of precision, coverage, and agreement. Then, we optimize it over all possible subsets of sizes to obtain a final \textit{committed} set. Though natural, this objective function is not submodular, and hence establishing its performance guarantees is not straightforward. We consider another objective function based on graph cut sub-modular function \cite{kothawade2021prism}. Our algorithm, \our, maximizes the objective function based on a greedy approach to obtain the final committed set from a large set of noisy rules. The algorithm works iteratively by selecting patterns having the highest incremental precision and marginal coverage with smaller conflicts over the unlabeled set to determine the top-\textit{k} rules.

Our setup favors the selection of a committed set in which the  LFs do not mutually contradict while maintaining overall good accuracy and precision. Most label aggregation models built with such committed sets should consequently yield lower noise in the labeling assigned to the unlabeled set. \our\ can be used to filter rules produced by any rule generation approach and is robust to any label aggregation approach because it only uses the characteristics of rules. To the best of our knowledge, such an approach of selecting a committed set has not been addressed in erstwhile approaches. 


We perform experiments with several ARI, pattern filtering, and label aggregation approaches as shown in Figure \ref{fig:flow1}. For generating candidate rules, we use approaches such as decision tree \cite{snuba}, classifier weights \cite{pryzant2022automatic}, and a modified version of \grasp\ \cite{shnarch2017Grasp}. Subsequently, we filter the large set of rules using the corresponding candidate generation approaches and our algorithm \our{}. Finally,  over the selected {\it committed set}  of rules, we leverage semi-supervised label aggregation algorithms, \textit{viz}, \spear~\cite{spear}, \astra~\cite{astra}, ImplyLoss \cite{awasthi2020learning} and Learning to Reweight \cite{l2r}.
We observe that our approach yields rules more refined than the other approaches by virtue of its analyzing (i) coverage and agreement over the unlabeled set and (ii) precision on the test set. 
We observe that rules filtered using \our{} perform better on label aggregation approaches, providing gains between 2 - 20\% across different datasets. It implies that the filtering of rules is a crucial element that was not explored earlier.

\section{Related Work}

ARI methods have largely focused on using repetitive structures or patterns in the tasks involving text documents, \textit{eg}, mentions of specific words or phrases \cite{varma2018snuba,shnarch2017Grasp}. Prior work relies on this observation to learn first-order logic rules as a composition of semantic role attributes \cite{sen2020learning} or syntactic grammar rules \cite{sahay2021rule}.
Recently, \citet{pryzant2022automatic} proposed a heuristic generation method that trains a classifier on the small labeled set and uses features corresponding to the \textit{k} largest values of weights as rules. Our proposed approach accounts for rule interdependence among a large set of generated heuristics and selects a useful subset of rules.
Our work is closest to interactive weak supervision~\cite{boecking2020interactive} which uses the active learning paradigm to select a useful set of final rules from a large rule set. However, our approach does not require the additional step of human annotations.

Prior work has emphasized on LFs defined by experts based on observations in a few instances from the dataset. Unsupervised approaches like \cite{bach2019snorkel} use a generative model to determine the correct probability for labels in accord with noisy and conflicting labels assigned by LFs. \citet{oishik} proposed a graphical model, CAGE, which extends to continuous LFs with scores obtained using cosine similarities of word vectors, TF-IDF score, the distance among entity pairs, \textit{etc.} while semi-supervised approaches additionally use a small labeled set to guide the discovery of LFs for classification \cite{abhishek2022spear}  and information extraction tasks \cite{singh2023eigen}. Recent methods \cite{maheshwari2022learning, sivasubramanian2023adaptive} proposed a bi-level optimization wherein parameters are optimized on the validation set in a semi-supervised manner. We use semi-supervised aggregation methods in our experiments.

\section{Background}

\subsection{Notations}
Let the feature space be $\mathcal{X}$ and the label space be $\mathcal{Y} \in \{1 \ldots K\}$ where $K$ is the number of classes. We have $M$ instances in the dataset out of which there are $N$ labeled instances denoted by set $\mathcal{L} = \{(x_i, y_i), i=1,2,\dots, N\}$ and $M-N$ unlabeled set of instance denoted by set $\mathcal{U} = \{x_i: i=N+1,N+2,\ldots, M\}$ where $M-N >> N$. The set of $m$ automatically induced rules is denoted by $ \mathcal{R} = (R_1, R_2 \ldots R_m)$, where $R_i: \mathcal{X}\rightarrow \mathcal{Y}\cup\{0\}$ for all $i=1,2,\ldots, m$. The label $0$  corresponds to an abstain decision by a rule. Each rule may abstain on a different set of instances. We denote the final set of $n$ filtered rules by $\mathcal{F} \subset \mathcal{R}$, where $n \leq m$.

\subsection{Rule Induction Methods}\label{sec:rule-methods}
We consider the following three methods for Automatic Rule Induction (ARI).

\textbf{Decision Tree}: \snuba~\cite{snuba} presented an ARI approach by using a small labeled set and fitting a decision tree over n-grams of the input sentence. Initially, rules are generated as a basic component of propositions on the labeled set. A proposition could be a word, a phrase, a lemma, or an abstraction such as part of a speech tag. Each composed rule is in the form of a decision stump (1-depth decision tree). \snuba\ is a three-step approach that (i) generates candidate rules using a labeled set, (ii) adds one rule based on the F1 score on the labeled set and Jacard score of the added rule, (iii) finds uncovered points or abstained points in the labeled set, and (iv) Removes the instances labeled in the labeled dataset by the added rule and repeat steps (i) - (iv) with updated labeled set. The process stops until the labeled set is completely covered or a limit on the number of iterations is reached.

\textbf{Classifier Weights} : 
\citet{pryzant2022automatic} trains a linear model classifier $C$ on the small labeled set. Suppose for $N$ instances in our dataset, each instance $x_i$ is denoted by its feature matrix $X_i$  of size $K$. The classifier model is $C(x_i) = \sigma(WX_i) $ where $W \in \mathcal{R}^{K\times N }$ is a weight matrix and $\sigma$ represents an element-wise sigmoid function. Then, it finds $P$ features corresponding to the largest weights in $W$ which is obtained by learning the classifier and creates one rule from each feature with $P$ largest weights.
If weight $w_{i,k}$ is assigned to the $i$-th feature, then they create a rule associated with the $i$-th feature and the $k$-th label. Here, rule filtering is limited to choosing $k$ rules having the largest weights in $W$.

\textbf{M-\grasp}: This is a modified version of \grasp\ \cite{shnarch2017Grasp} for automatically extracting the patterns that characterize subtle linguistic phenomena. M-\grasp\ augments each term of the text with multiple attributes such as lemma, hypernyms, NER, POS tags {\em{etc.}} to extract the rich set of generalizable patterns. The algorithm expects a considerable sized set of positive and negative samples to extract the discriminative patterns. In contrast to \grasp\, we have a small labeled set and a large unlabeled set with multiple classes. To make the core algorithm pertinent to our setting, we make two important modifications to  \grasp. (i) The original \grasp\ algorithm uses the entire labeled set to generate an initial candidate set of patterns of length 1 but in  M-\grasp\ we also employ the unlabeled set to generate these patterns. However, during the iterative process,  we filter patterns using the information gain measure on the labeled set. (ii) While the original \grasp\ algorithm assumes binary classes,  we extend the algorithm for the multi-class setting as well. For generating new patterns in M-\grasp\ we follow the same iterative process of the original \grasp\ algorithm other than the above-mentioned changes. 

\section{Problem Setup}
We begin by defining key metrics used throughout the experiments. a) \textbf{Precision} of a rule on a labeled set is the ratio of the number of correctly assigned labels over the total assigned labels,  b) \textbf{coverage} of a rule is the percentage of points covered over unlabeled set, c) \textbf{conflicts} between rules $R_j$ and $R_i, j\neq i$, is the percentage of points that are assigned different labels by rule $R_i$ and $R_j$ and d) \textbf{agreement} between rules $R_i$ and $R_j$ is defined as the percentage of points that are assigned same labels by rules $R_i$ and $R_j$.
Note that, conflicts and agreements are related as both rules on a particular instance will either conflict or agree.
We consider different submodular functions reflecting precision, coverage, and conflicts of the rules as objectives. We also explore their combinations as objective functions. 

\subsection{Precision Coverage Agreement Objective - $f_{PCA}$}

Given a labeled set $\mathcal{L}$ and a candidate set of rules $\mathcal{R}$, our aim is to find a final set of rules $\mathcal{F}\subseteq \mathcal{R}$, which has high precision and coverage but fewer conflicts. 
Initially, we propose the following function:
\begin{multline}
 f_{PCA}(\mathcal{F})=w *\alpha(\mathcal{F})/|\mathcal{F}|+(1-w) * \beta(\mathcal{F})\\ + \gamma*\mu(\mathcal{F})
 \label{eq:pc}
 \end{multline}
 where $\alpha(\mathcal{F})=\sum_{R_i\in \mathcal{F}}\text{Precision}(R_i)$, $\beta(\mathcal{F})$ = coverage($\mathcal{F}$) and $\mu(\mathcal{F})$ as the percentage of instances over which all the rules in $\mathcal{F}$ provide non-conflicting labels. Given a maximum number of rules $k$, we define our objective as $\max_{|\mathcal{F}|\leq k} f_{PCA}(\mathcal{F})$.
 We observe that  \eqref{eq:pc} does not satisfy the submodular properties of the function \cite{wei2015submodularity} (see Appendix \ref{ex: 1}). Hence, we cannot secure theoretical guarantees for choosing the optimal rule subset. Since submodularity provides theoretical guarantees for the optimization problem, we substituted the objective as $f_{GC}$ described in the next subsection. However, $f_{PCA}$ still provides interpretable rules and competitive results. We perform a qualitative analysis of rules against the other variant in the Appendix \ref{sec: Interesting obs}.
Below, we describe the algorithm for $f_{PCA}$ in Algorithm \ref{alg:algo-fpca}. We define $Cov$L as the function that outputs the coverage on the labeled set. Then, we add that rule to the committed set which maximizes the contribution i.e. $f_{PCA}\{\mathcal{F}\cup \{r\}\}-f_{PCA}\{\mathcal{F}\}$ as in line 5 of the algorithm.

\begin{algorithm}
\caption{\our{} Precision Coverage - $f_{PCA}$}\label{alg:algo-fpca}
\begin{algorithmic}[1]
\State \textbf{Input:} Candidate set of rules $\mathcal{R}$, Labeled set $\mathcal{L}$, Unlabeled set $\mathcal{U}$, final set of rules $\mathcal{F}$,  Hyperparameters : $w$, $\gamma$, $k$
\State Initialize $\mathcal{F}$ = $ argmax_{i}(f_{PCA}(R_i)) \hspace{0.2cm} \forall i \in \mathcal{R}$
\While {$CovL(\mathcal{F}) < 1.0$ and $|\mathcal{F}|< k$}

\State $r^{*}\gets argmax_{r\in \mathcal{R}-\mathcal{F}}(f_{PCA}\{\mathcal{F}\cup\{r\}\}-f_{PCA}\{\mathcal{F}\})$

\State $\mathcal{F} \gets \mathcal{F}\cup\{r^{*}\}$
\EndWhile
\State \textbf{Output:} $\mathcal{F}$
\end{algorithmic}
\end{algorithm}
\textbf{Termination Condition:} In this variant, we have used a stopping criteria as if every instance in the labeled set got covered then we will stop. This condition states that if the labeled set is covered then we assume that rules in the committed set are diverse enough to label the unlabeled data.

\subsection{Graph-Cut Submodular Objective - $f_{GC}$}

Let $\Omega $ be a set of elements. A function $f: 2^{\Omega}\rightarrow \mathbb{R}$ is said to be submodular if it satisfies the property of diminishing returns i.e. for every $A\subseteq B \subseteq \Omega$ and $j\notin B$, $f(A\cup\{j\})-f(A)\geq f(B\cup\{j\})-f(B)$. A greedy algorithm provides an $\mathcal{O}(1)$-approximation to the optimal solution. Due to this algorithmic property, subset selection with submodular objective functions has found several applications in text summarization \cite{yao2017recent}, video summarization \cite{kaushal2019demystifying,gygli2015video}, training speed up \cite{kaushal2019learning}, active learning \cite{settles2009active}, {\em{etc.}}
Also, submodular functions have found wider acceptance due to their ability to naturally model the notion of representativeness, diversity, and coverage. Hence, we pose our rule selection problem within a submodular subset selection framework.


\citet{kothawade2021prism} presents a wide array of submodular functions and their variants such as mutual information and their conditional gain counterparts. From that wide spectrum, we choose graph-cut (GC) that selects both representative and diverse instances from the ground set. We consider the candidate set of rules as the ground set and obtain a set of diverse and representative rules. For any $\mathcal{F} \subset \mathcal{R}$, the GC function is defined as follows:
\begin{equation}\label{eq:gc}
    f_{GC}(\mathcal{F})=\sum_{i\in \mathcal{R}, j\in \mathcal{F}} s_{ij}-\lambda \sum_{i,j\in \mathcal{F}}s_{ij}
\end{equation}
where $\lambda \in [0,1]$ governs the trade-off between diversity and representation. Higher $\lambda $ selects a diverse set of rules. $s_{ij}$ is the similarity score for rule pair $R_i$ and $R_j$.
We propose a similarity score $s_{ij}$ as:
\begin{equation}\label{eq:worth}
    s_{ij} = \alpha(R_i)+\alpha(R_j)+w*\beta(\{R_i, R_j\})\\+\gamma*\mu(R_i,R_j)
\end{equation}
where $\alpha(R_i)=\text{Precision}(R_i)$ and $\beta(\{R_i, R_j\})$ is the coverage function that gives the coverage of both the rules $R_i$ and $R_j$, i.e., fraction of the unlabeled set that is labeled by at least one of the rule.  $\mu(R_i,R_j)$ is the agreement between rule $R_i$ and $R_j$. It is defined as the fraction of the unlabeled instances on which both rules provide the same labels.   
In Eq.~\ref{eq:worth}, $w$ and $\gamma$ denotes the weights of $\beta$ and $\mu$ respectively. While $w$ regulates the weight given to the coverage component, $\gamma$ regulates the weight given to the agreement component. The range of values of $w$ and $\gamma$ are chosen by carefully analyzing the statistics related to the coverage and precision.

\noindent\textbf{Tuning $w$ and $\gamma$}  We tune $w$ and $\gamma$ based on the validation dataset. We observed the quality of rules when $w$ is $1\leq w \leq 10$ and  $\gamma$ is $0\leq \gamma \leq 1$. We found that when $w$ is between 2 and 4, coverage component $\beta$ has weightage comparable to the precision components. Intuitively, both components contribute equally while producing the final committed rule set. $\gamma$ refers to the weighing factor for agreement between rule $R_i$ and $R_j$. We found the best rule set when $\gamma$ is between 0.2 and 0.5. The high agreement between various rules compensates for the lower values of $\gamma$.

In the GC function, $\lambda\in [0,1]$ governs the trade-off between representation and diversity and we need a committed set that is diverse enough. In our experiments, we set $\lambda$ to 0.7 so that the final candidate rules are diverse in nature. GC is a non-monotone submodular for $\lambda > 0.5$, hence in our case, $f_{GC}$ is a non-monotone submodular function. 

Given a cardinality budget constraint on the number of rules $k$, our  objective function is,
$\max_{|\mathcal{F}|\leq k} f_{GC}(\mathcal{F})$
which is a submodular function \cite{kothawade2021prism}. We use a greedy algorithm for maximizing this function. We greedily choose the rule that maximizes the marginal utility i.e.
$\textit{argmax}_{i\in\{\mathcal{R}-\mathcal{F}\}} f_{GC}(\mathcal{F}\cup \{i\})-f_{GC}(\mathcal{F})$.
The greedy algorithm begins with an empty set and then iteratively adds a rule from the candidate set to the committed set by maximizing the marginal gain in every iteration until the budget constraint is met. The pseudo-code of our approach is given in Algorithm \ref{algo:FAIR}.
For a candidate set of rules $\mathcal{R}$ induced from ARI approaches (Section \ref{sec:rule-methods}), we compute precision on $\mathcal{L}$ using $findprecision$ and coverage over $\mathcal{U}$ using $findcoverage$ for each rule $R_i$. We calculate the agreement between two rules $R_i, R_j$ using the $findagreement$ function. Then, we compute $s_{ij}$, $(i,j)^{th}$ entry of matrix $S$ defined as in line 8. Finally, we find the set of committed rules $\mathcal{F}$ using Eq. \ref{eq:gc} for a pre-specified budget of $k$ rules.
Note: In the following sections \our{} refers to the GC variant of \our{}, unless otherwise stated.
\begin{algorithm}
\caption{\our{} Graph cut}\label{algo:FAIR}

\begin{algorithmic}[1]
\State \textbf{Input:} Candidate set of rules $\mathcal{R}$, Labeled set $\mathcal{L}$, Unlabeled set $\mathcal{U}$, final set of rules $\mathcal{F}$, $w$, maximum number of LFs $k$
\State Initialize $ S  = [0]_{|\mathcal{R}|\times |\mathcal{R}|} $

\State $\alpha(R_i) \gets findprecision(R_i)$
\State $\beta(R_i+R_j) \gets findcoverage(\{ R_i\cup R_j\})$
\State $\mu(R_i, R_j) \gets findagreement(R_i, R_j)$

\State $S(i,j) \gets \alpha(R_i)+\alpha(R_j)+w*\beta(R_i+R_j)+\gamma*\mu(R_i, R_j) \text{  }\forall i \neq j$
\State $\mathcal{F} \gets GraphCut(|R|, S, k)$

\State \textbf{Output:} $\mathcal{F}$
\end{algorithmic}

\end{algorithm}

\section{Experiments}
We select five text classification datasets and compare rules induced by \our{}  with the rules generated by three ARI approaches ({\em c.f.}, Section ~\ref{sec:rule-methods}). We measure the efficacy of these rules by aggregating the assigned labels using four semi-supervised label aggregation approaches. It takes around approximately 2 - 6 GPU hours to complete all experiments except SST which takes around 12 hours to complete. We performed all our experiments on a single Nvidia RTX 2070.

\noindent\textbf{Datasets} : (1) \textbf{TREC} \cite{trec} is a multi-class question classification dataset consisting of open-domain, fact-based questions divided into broad semantic categories. The dataset has the following six labels: \textit{Abbreviation, Description, Entities, Human, Locations} and \textit{Numeric values}.  (2) \textbf{YouTube Comment  Classification} \cite{youtube} is a spam comment classification dataset. (3) \textbf{IMDB Genre Classification} \footnote{http://www.imdb.com/datasets} is a binary movie-genre classification dataset from a movie plot summary. The labels are \textit{romantic} and \textit{action}. (4) \textbf{SMS Spam Classification} \cite{sms} is a binary classification dataset to classify a given sms as \textit{spam} or \textit{not spam}. (5) \textbf{Stanford Sentiment Treebank (SST)} \cite{sst5} is a collection of written or spoken texts with fully labeled parse trees for a complete analysis of the compositional effects of sentiment in language. The output labels for this dataset are \textit{negative, somewhat negative, neutral, somewhat positive} and \textit{positive}.




\begin{figure*}
    \centering
    \includegraphics[width=8.5cm]{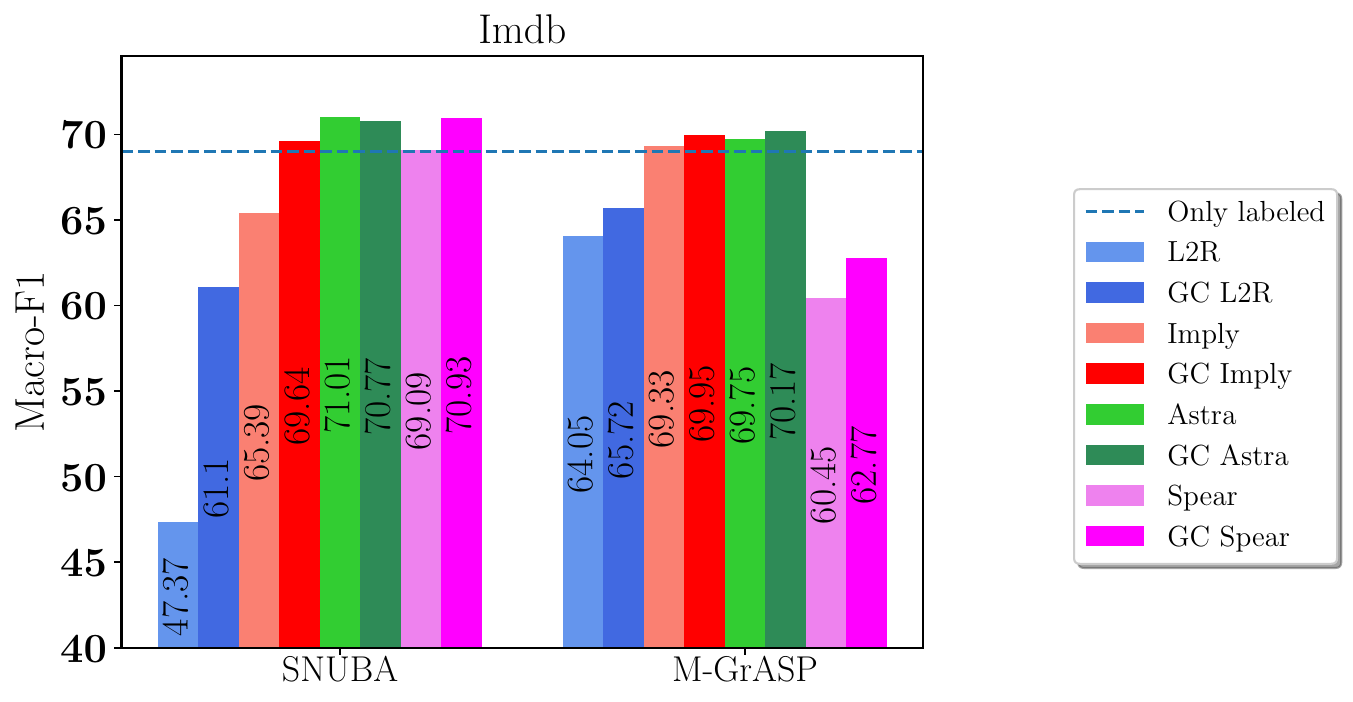}
    \includegraphics[width = 6.0cm]{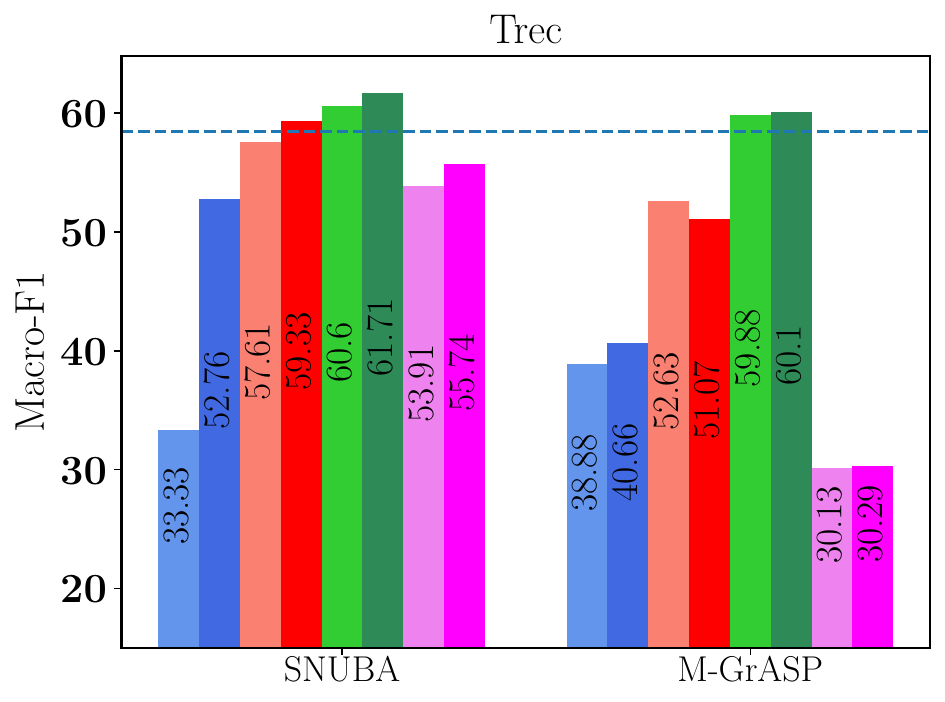}
    \caption{ Results for IMDB and TREC dataset for \our\ GC against \snuba\ and M-\grasp.}
    \label{fig:res2}
\end{figure*}

\begin{figure*}
    \centering
    \includegraphics[width=8.5cm]{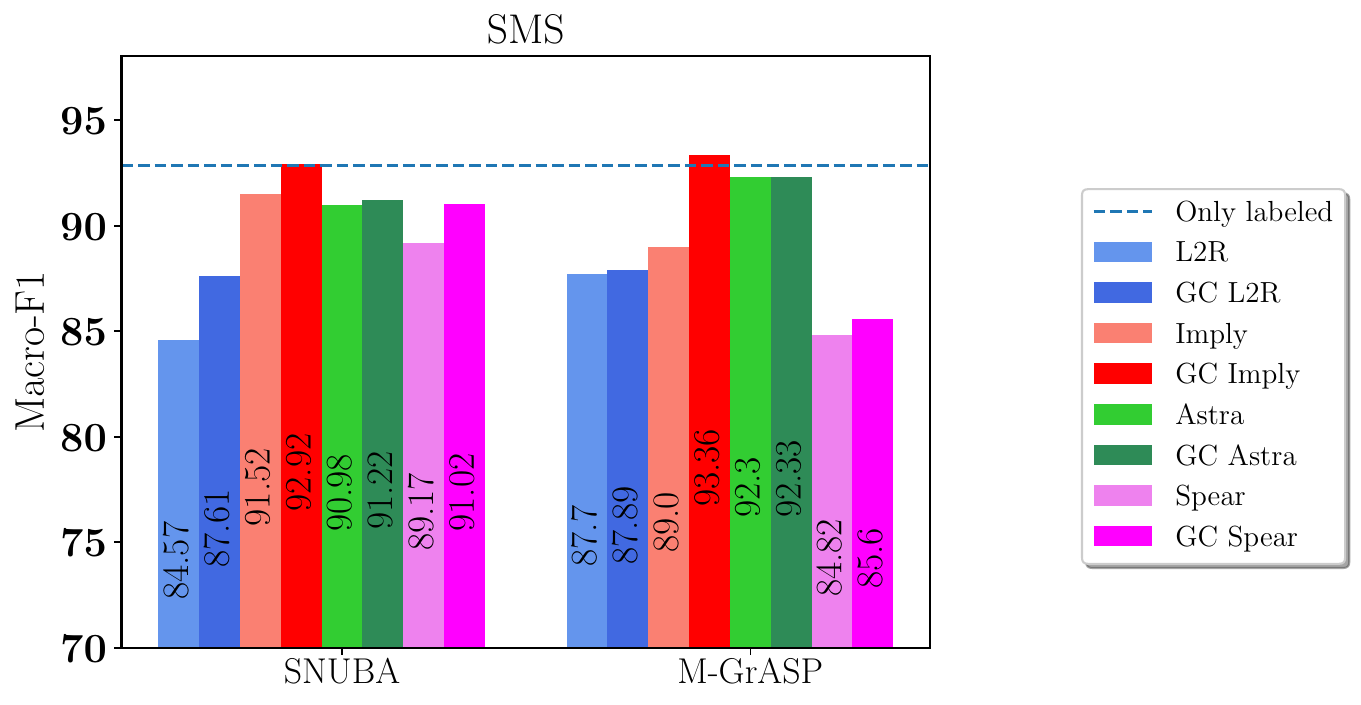}
    \includegraphics[width = 6.0cm]{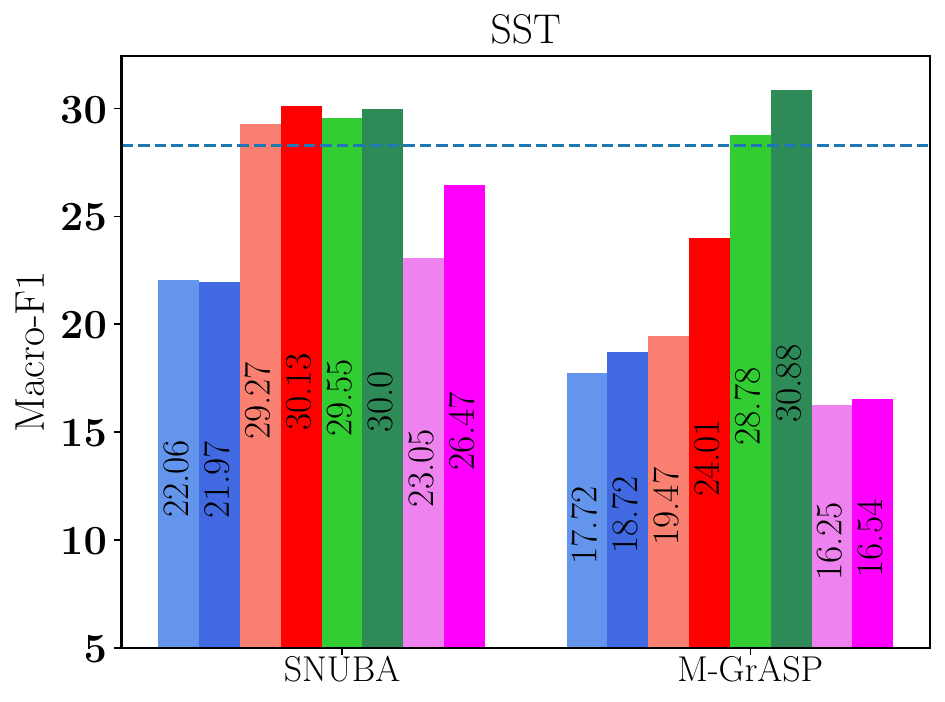}
    \caption{Comparison of \our{} against \snuba\ and M-\grasp\ filtering over different label aggregation approaches, GC is \our{} GraphCut. The size of the final committed set is the same across all ARI approaches.
}
\label{fig:res1}
\end{figure*}

\subsection{Label aggregation methods}
We use a small labeled set $\mathcal{L}$ to automatically induce rules using different ARIs. We form an Only-$\mathcal{L}$ baseline, where we train a supervised classifier on the small labeled set $\mathcal{L}$ using standard cross-entropy loss. The network architecture for Only-$\mathcal{L}$ is similar to \citet{awasthi2020learning}.
We use the following semi-supervised label aggregation approaches:

\noindent \textbf{Learn to Reweight (L2R)} \cite{l2r}: This approach uses the noisy labels provided by rules as well and trains the classifier by meta-learning to reweight the loss on noisily labeled instances, and for performing this step clean labeled dataset $\mathcal{L}$ is utilized.

\noindent\textbf{Imply Loss} \cite{awasthi2020learning}: It uses additional information in the form of labeled rule exemplars jointly denoises rules via latent coverage variables, and trains a model on soft implication loss over coverage and label variables. 

\noindent \textbf{\spear} \cite{spear} is a semi-supervised paradigm that jointly learns the parameters over features and labeling functions (rules) in a semi-supervised manner. It jointly learns a parameterized graphical model and a classifier model.

\noindent \textbf{\astra} \cite{astra} is a weak-supervision framework that uses all available data both labeled and unlabeled set in an iterative self-training framework. It trains a student model on unlabeled data that considers contextualized representations and predicts pseudo-labels for instances not covered by rules. Thereafter, it learns a rule attention network that learns to aggregate pseudo-labels assigned by the student model in conjunction with noisy labels assigned by rules. An iterative student-teacher model is trained with a semi-supervised objective.

\subsection{Experimental Setting}
We use 10\% of the dataset as a labeled set to generate rules for our model. The 10\% labeled set is split equally for the label aggregation stage. We reserve 5\% of the total corpus as the labeled set and 5\% as a validation set while the rest of the set is unlabeled. We performed a final evaluation on 500 instances for each dataset (refer Table \ref{tab: dataset} in Appendix).
The remaining portion of the dataset was left unlabeled. We use \snuba\ with raw count-based features with a decision tree to generate a candidate rule set. M-\grasp\ uses lemma, hypernym, text, and sentiment-based attributes to generate the rule set. The classifier weight approach uses logistic regression to train a classifier model for finding the top-\textit{k} weights and associated features with these weights as rules.
For label aggregation methods, we follow the same hyper-parameters as provided in the respective codebases. The features set is the same as followed in the \spear. It yields the best performance on the combination of first, third, fourth, and sixth loss combination. On all datasets, macro-F1 is employed as the evaluation criterion. 
Performance numbers for each experiment are obtained by averaging over five independent runs, each having a different random initialization.\\
\textbf{Rule-level analysis} 
We also performed analysis on the rules retrieved from \our{} and compared them with other filtering approaches. 
By qualitatively analyzing the rules, we found that \our{} returns a more interpretable and diverse collection of rules. By analyzing rules provided in the Appendix, we observe that \our{} is not prejudiced against any particular class. We see that different filtering techniques frequently favor a single class more than others. The diversity of the rules aids in covering more instances, leading to higher-end-model performance. One particular example from the SMS dataset could be \textit{"www"} is a highly precise rule denoting the \textit{spam} class was missed by the classifier's filtering method but covered by \our{}.
Other than qualitative analysis, we also provide statistics of rules such as coverage, agreement, and precision on the test set of the committed set of rules. We compare them against different filtering approaches.  
Due to the paucity of space, we present a full discussion in the Appendix section.

\begin{figure}
\includegraphics[scale=.45]{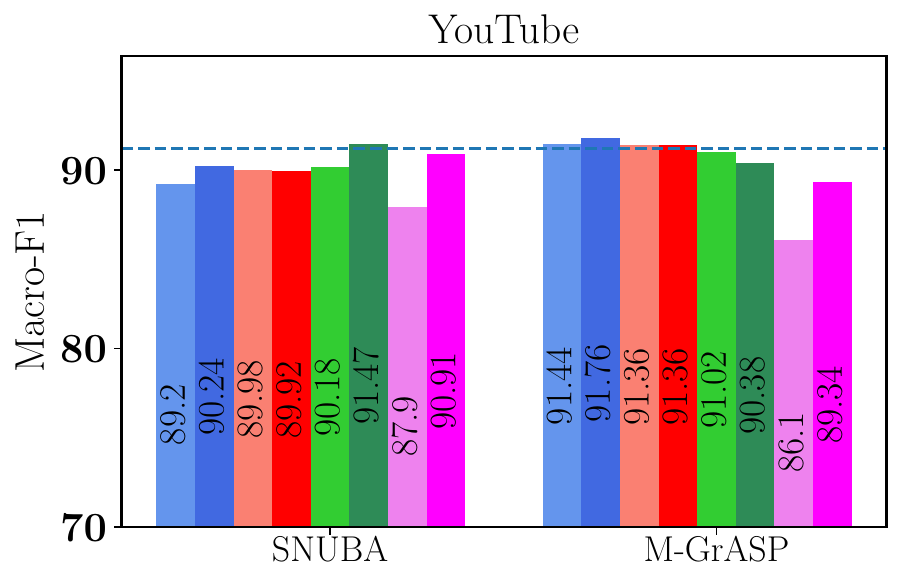}
\caption{Results on YouTube dataset for \our\ GC}\label{fig:res9}
\end{figure}

\begin{figure*}
    \centering
    \includegraphics[width=8.5cm]{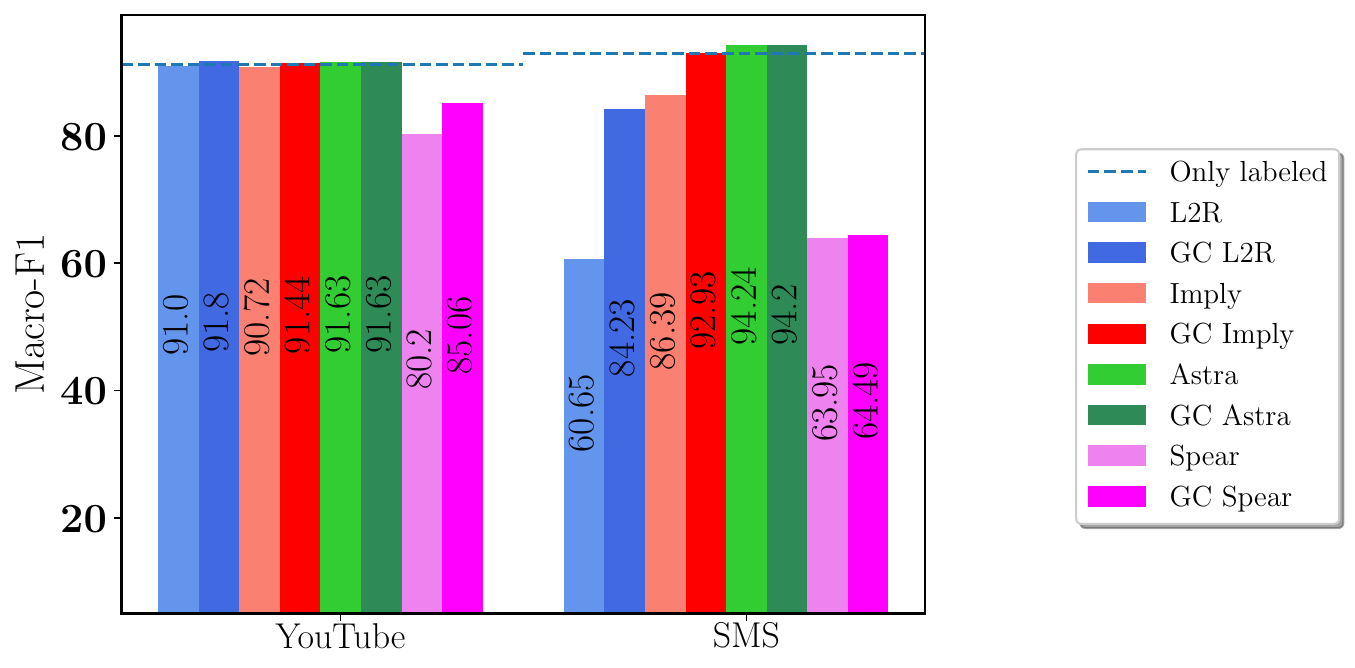}
    \includegraphics[width=6.0cm]{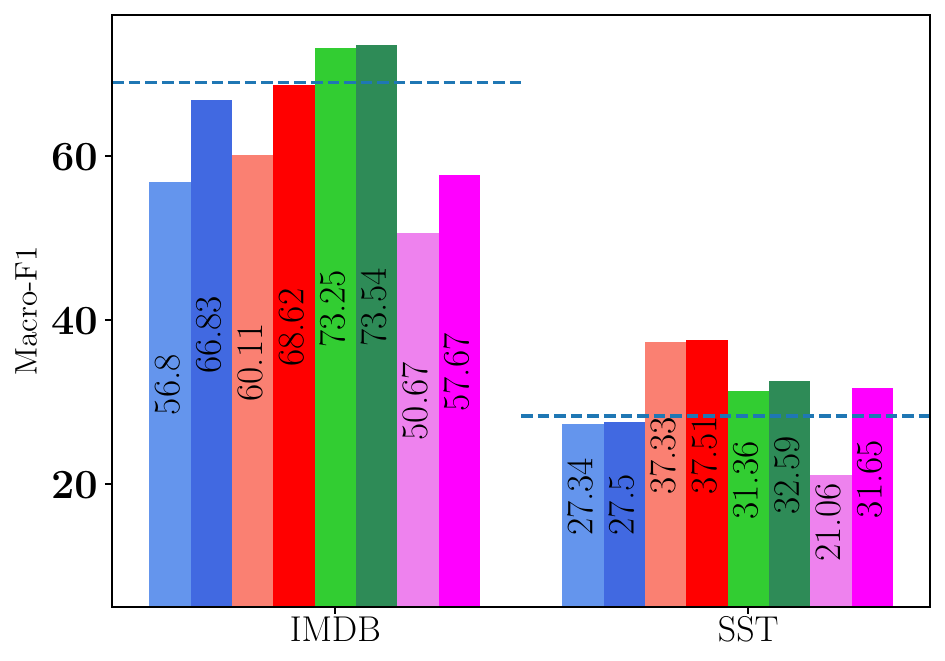}
    \caption{Comparison of \our\ GC against Classifier weights on different datasets.}
    \label{fig:res6}
\end{figure*}

\subsection{Results}
In Figure \ref{fig:res2}, \ref{fig:res1}, \ref{fig:res9} and \ref{fig:res6}, we present results for the four label aggregation methods for various datasets and rule filtering approaches (\snuba\, \grasp\, Classifier weights and \our{}). For the SMS dataset, \our{} achieves better performance than rules induced by \snuba, M-\grasp, and Classifier weights for all label aggregation approaches. We achieve maximum gains on L2R by up to 19 points and up to 5-point gains over ImplyLoss. Similarly, on \astra\ maximum gains are a bit lesser of about 2 points, and on \spear\ maximum gains are 10 points. On the SST dataset, we observe a performance drop on L2R, however, gains are consistent for other approaches. This could be possible due to the aggregation scheme of L2R. 
\our{} consistently outperforms all the other filtering approaches across all label aggregation approaches and across all the datasets with a significant margin. 

\noindent\textbf{On \snuba} In comparison to \snuba\, we observe better end-model performance using rules filtered by \our{}. In each iteration, \snuba\ reduces the size of its candidate set thus missing out on important rules. \our{} produces the committed set of rules in a single iteration and chooses a diverse set of rules resulting in rules having higher coverage and competitive agreement.

\noindent\textbf{On Classifier weights} Since classifier weights choose top-$K$ rules according to weights of the features, it does not model the coverage of a rule on an unlabeled dataset unlike \our{}. Further, it does not explicitly model the agreement between rules resulting in conflicting rules (See Table \ref{tab:stats-cw}).

\noindent\textbf{On M-\grasp}: M-\grasp\ chooses the committed set according to the information gain of each rule. However, it does not model agreement as well as the diversity of rules in the committed set. In Table \ref{tab:stats-grasp}, we observe consistently better coverage and precision with a competitive agreement resulting in better performance for most of the datasets.\\
\textbf{Observations on \astra}: \astra\ uses self-supervision and uses a weighted sum of labels provided by rules and the student(classifier) model. While tuning the hyperparameter, we observed that the student model has a huge impact on the labels provided to unlabeled instances reducing the impact of rules on the final number. Even though rules have higher conflicts and low coverage in the committed set, \astra\ yields comparable numbers, unlike other aggregation approaches. This justifies lower gains over the \astra\ approach and even hits the YouTube dataset with M-\grasp\ rule generation.

\begin{figure*}
    \centering
    \includegraphics[width=8.3cm]{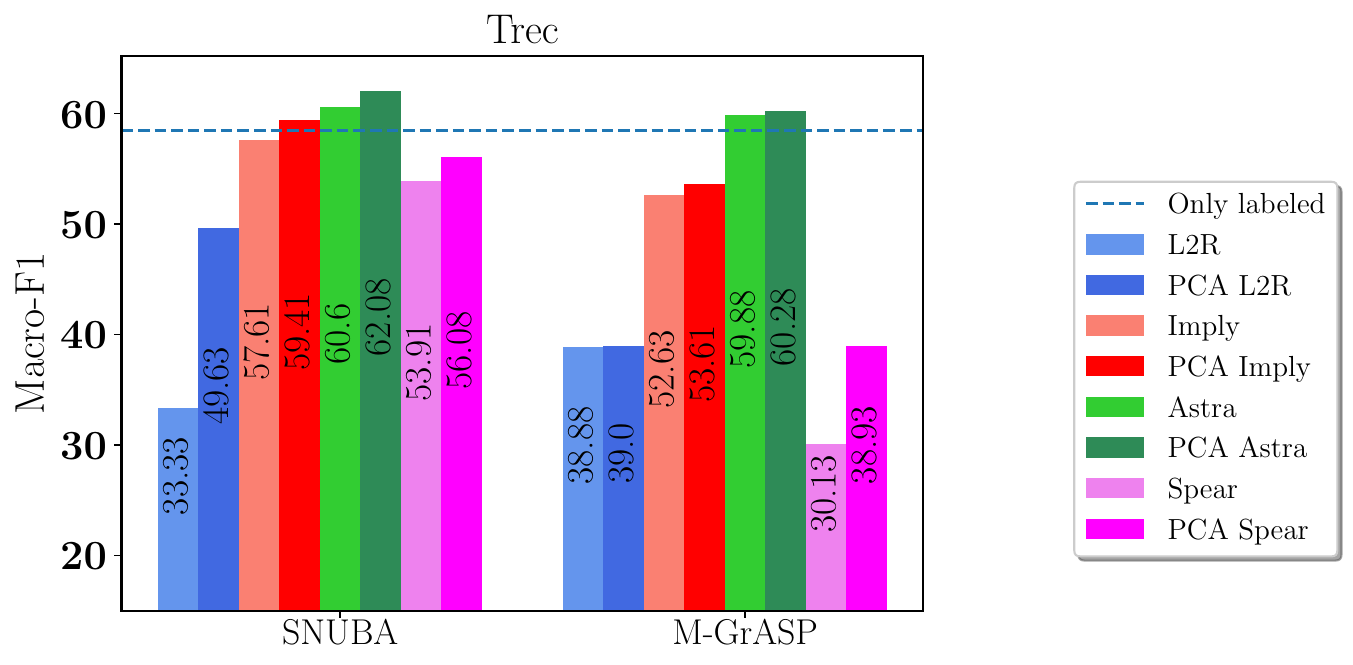}
    \includegraphics[width = 6.3cm]{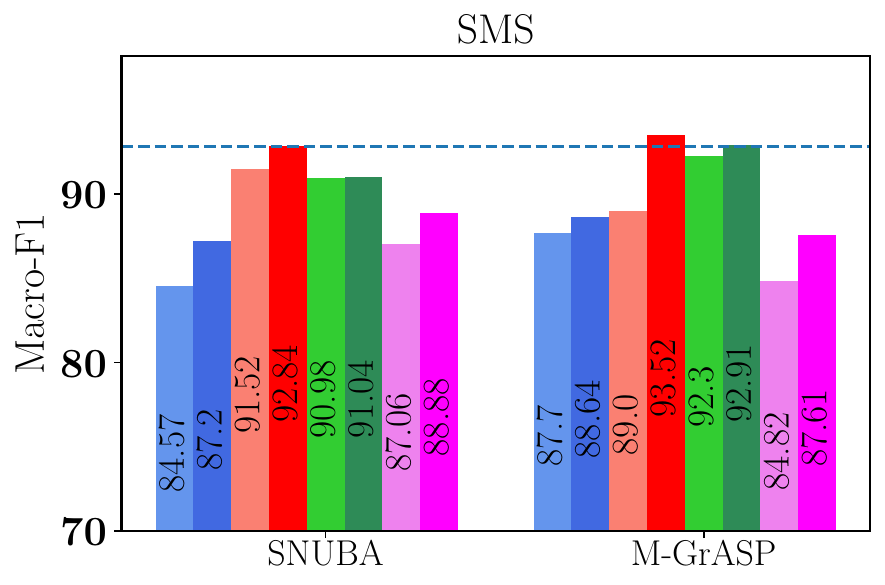}
    \caption{Comparison of Macro F1 score of \snuba\ and M-\grasp\ on different aggregation approaches for TREC dataset and SMS dataset.}
    \label{fig:res5}
\end{figure*}

\begin{figure*}
    \centering
    \includegraphics[width=8.3cm]{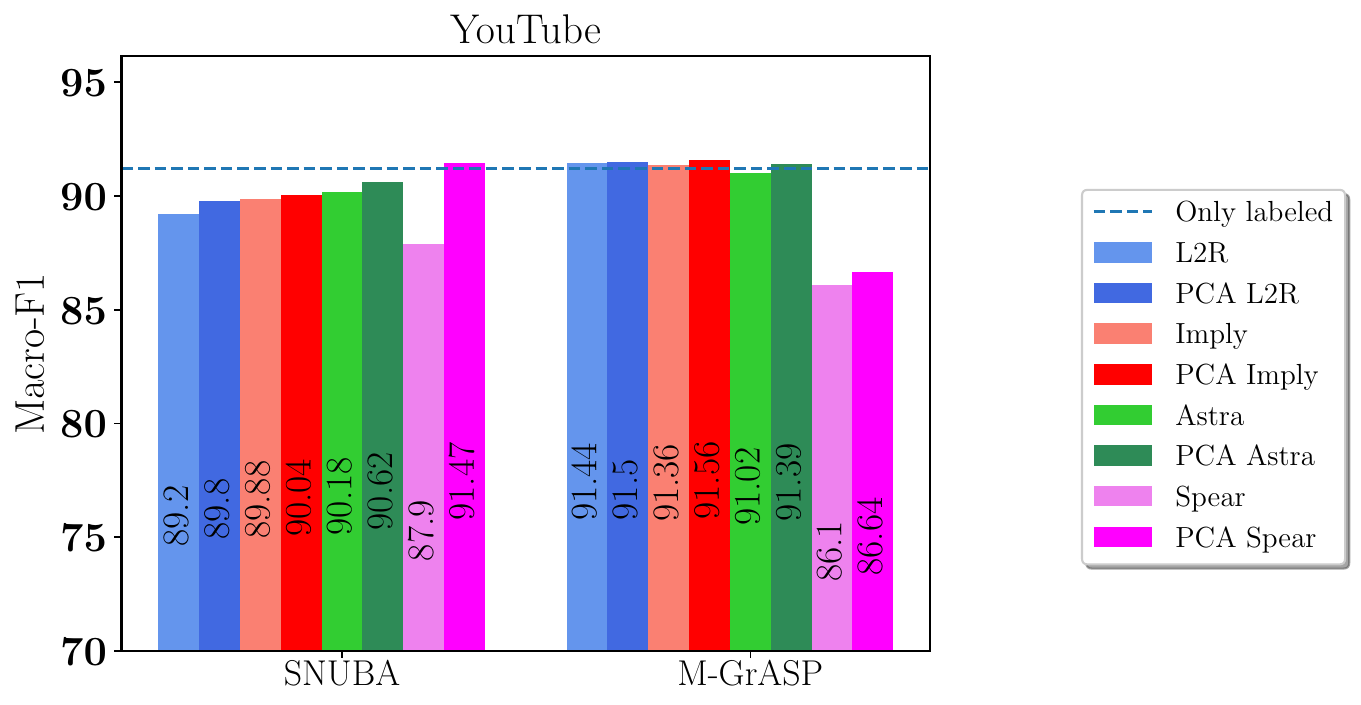}
    \includegraphics[width = 6.5cm]{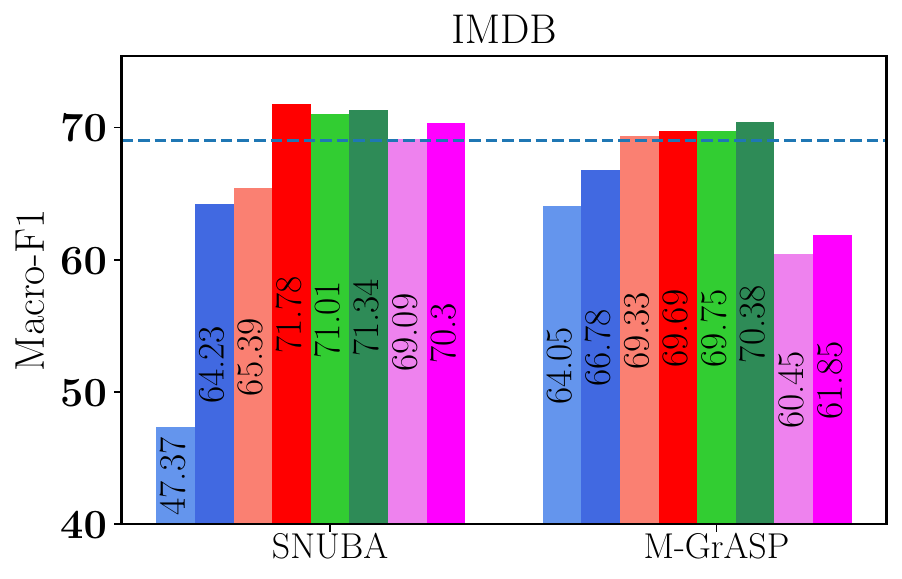}
    \caption{Comparison of Macro F1 score of \snuba\ and M-\grasp\ on different aggregation approaches for YouTube and IMDB dataset.}
    \label{fig:res4}
\end{figure*}

\subsection{Significance test}
We employ the Wilcoxon signed-rank test \cite{wilcoxon1992individual} for statistical significance test. We chose the null hypothesis as there is no significant difference between \our{} and other filtering approaches and we successfully rejected it over all datasets. We had a value of n = 20 as there are 5 datasets and 4 aggregation approaches. \our{} is statistically significant at $p< 0.05$ than all other filtering methods across all label aggregation approaches. These results suggest that \our{} is robust to any label aggregation as well as any rule induction approach.
$p$ and $z$- values are reported in table \ref{tab:Significance tests}.
\subsection{Comparison between $f_{GC}$ and $f_{PCA}$}\label{sec: Interesting obs}
Though theoretically not very promising, $f_{PCA}$ implemented in algorithm 1, has reasonable interpretation and comparable performance. This section discusses the results obtained by implementing algorithm 1 on the datasets. In Figure \ref{fig:res5},
\ref{fig:res4}, and \ref{fig:res10}, we demonstrate the results for $f_{PCA}$  and compare them against previous filtering approaches. We use the same data split as for \our{} GC. The numbers reported here are averaged over five independent runs over random generalizations. We tuned $w$ and $\gamma$ over a validation set. 
We note that we are getting constant gains using \our{} PCA, but the gains are small as compared to \our{} GC in most cases. In most of the datasets, initially, the rules chosen from the candidate set were the same for \our{} GC as well as \our{} PCA since the objective function performs similarly until the cardinality of the committed set is small. As more rules were added, we observed different rules being added in the committed set of rules as \our{} GC chooses more diverse rules using $\lambda$ parameter. One such example could be found in Table \ref{tab: QA1} in the Appendix. In that table, each italic entry is a rule which assigns a weak label to an instance. Observe that GC rules make more sense than CW rules as \textit{shuffle} can be reduced from the committed set of CW rules as it is not a likely word in a spam or ham comment. If we compare GC rules and PCA rules, we can observe that GC rules are more diverse covering rules from different classes. From Table \ref{tab: QA1}, we observe PCA rules have only one rule in class \textit{ham} while GC has three proving it to be more diverse than PCA.

\section{Conclusion}
We propose \our{}, a rule-filtering approach that selects a useful subset of rules from a given large candidate set of rules by leveraging implicit interdependencies among the rules. We introduce an objective function that maximizes precision, coverage, and agreement among rules and augments the function by designing a submodular function providing convergence guarantees. We conduct extensive experiments and demonstrate the importance of selecting high-quality and diverse rules with very few labeled instances. We qualitatively analyze the rules generated by \our{} and existing approaches. Further, we show that \our{} outperforms existing filtering approaches in terms of end-model performance using different label aggregation methods which makes \our{} robust to different aggregation as well as rule generation approaches.

\section{Limitations}
A key limitation is the performance of our approach on rule sets that are more noisier than current datasets.
Our benchmark rule-filtering methods rely on generating and filtering via the same approach. An enhanced benchmark could encompass rule generation through one approach and subsequent filtering through a different method. The ARI approach is linked to the size of the labeled set. With an increase in the size of this set, the time required for rule generation and the rule-filtering method also correspondingly increases.

\section*{Acknowledgements}
We thank Vineeth Dorna for his contributions during initial discussions of this work. Divya Jyoti Bajpai is supported by the Prime Minister’s Research Fellowship. Ayush Maheshwari was supported by a fellowship from the Ekal Foundation during his PhD at IIT Bombay. 
Manjesh K. Hanawal thanks funding support from SERB, Govt. of India, through the Core Research Grant (CRG/2022/008807) and MATRICS grant (MTR/2021/000645), and DST-Inria Targeted Programme. Ganesh Ramakrishnan is grateful to the National Language Translation Mission (NLTM): Bhashini project by Government of India and IIT Bombay Institute Chair Professorship for their support and sponsorship.

\bibliography{EACL2024/ref}

\begin{thebibliography}{33}
\expandafter\ifx\csname natexlab\endcsname\relax\def\natexlab#1{#1}\fi

\bibitem[{Abhishek et~al.(2022)Abhishek, Ingole, Laturia, Dorna, Maheshwari,
  Ramakrishnan, and Iyer}]{abhishek2022spear}
Guttu Abhishek, Harshad Ingole, Parth Laturia, Vineeth Dorna, Ayush Maheshwari,
  Ganesh Ramakrishnan, and Rishabh Iyer. 2022.
\newblock Spear: Semi-supervised data programming in python.
\newblock In \emph{Proceedings of the The 2022 Conference on Empirical Methods
  in Natural Language Processing: System Demonstrations}, pages 121--127.

\bibitem[{Alberto et~al.(2015)Alberto, Lochter, and Almeida}]{youtube}
T{\'u}lio~C Alberto, Johannes~V Lochter, and Tiago~A Almeida. 2015.
\newblock Tubespam: Comment spam filtering on youtube.
\newblock In \emph{2015 IEEE 14th international conference on machine learning
  and applications (ICMLA)}, pages 138--143. IEEE.

\bibitem[{Almeida et~al.(2011)Almeida, Hidalgo, and Yamakami}]{sms}
Tiago~A Almeida, Jos{\'e} Mar{\'\i}a~G Hidalgo, and Akebo Yamakami. 2011.
\newblock Contributions to the study of sms spam filtering: new collection and
  results.
\newblock In \emph{Proceedings of the 11th ACM symposium on Document
  engineering}, pages 259--262.

\bibitem[{Asano et~al.(2019)Asano, Rupprecht, and Vedaldi}]{asano2019critical}
YM~Asano, C~Rupprecht, and A~Vedaldi. 2019.
\newblock A critical analysis of self-supervision, or what we can learn from a
  single image.
\newblock In \emph{International Conference on Learning Representations}.

\bibitem[{Awasthi et~al.(2020)Awasthi, Ghosh, Goyal, and
  Sarawagi}]{awasthi2020learning}
Abhijeet Awasthi, Sabyasachi Ghosh, Rasna Goyal, and Sunita Sarawagi. 2020.
\newblock \href {https://openreview.net/forum?id=SkeuexBtDr} {Learning from
  rules generalizing labeled exemplars}.
\newblock In \emph{8th International Conference on Learning Representations,
  {ICLR} 2020, Addis Ababa, Ethiopia, April 26-30, 2020}. OpenReview.net.

\bibitem[{Bach et~al.(2019)Bach, Rodriguez, Liu, Luo, Shao, Xia, Sen, Ratner,
  Hancock, Alborzi et~al.}]{bach2019snorkel}
Stephen~H Bach, Daniel Rodriguez, Yintao Liu, Chong Luo, Haidong Shao,
  Cassandra Xia, Souvik Sen, Alex Ratner, Braden Hancock, Houman Alborzi,
  et~al. 2019.
\newblock Snorkel drybell: A case study in deploying weak supervision at
  industrial scale.
\newblock In \emph{Proceedings of the 2019 International Conference on
  Management of Data}, pages 362--375.

\bibitem[{Boecking et~al.(2020)Boecking, Neiswanger, Xing, and
  Dubrawski}]{boecking2020interactive}
Benedikt Boecking, Willie Neiswanger, Eric Xing, and Artur Dubrawski. 2020.
\newblock Interactive weak supervision: Learning useful heuristics for data
  labeling.
\newblock \emph{arXiv preprint arXiv:2012.06046}.

\bibitem[{Brabham(2013)}]{brabham2013crowdsourcing}
Daren~C Brabham. 2013.
\newblock \emph{Crowdsourcing}.
\newblock Mit Press.

\bibitem[{Chatterjee et~al.(2020)Chatterjee, Ramakrishnan, and
  Sarawagi}]{oishik}
Oishik Chatterjee, Ganesh Ramakrishnan, and Sunita Sarawagi. 2020.
\newblock Robust data programming with precision-guided labeling functions.
\newblock In \emph{AAAI}.

\bibitem[{Gygli et~al.(2015)Gygli, Grabner, and Van~Gool}]{gygli2015video}
Michael Gygli, Helmut Grabner, and Luc Van~Gool. 2015.
\newblock Video summarization by learning submodular mixtures of objectives.
\newblock In \emph{Proceedings of the IEEE conference on computer vision and
  pattern recognition}, pages 3090--3098.

\bibitem[{Karamanolakis et~al.(2021)Karamanolakis, Mukherjee, Zheng, and
  Hassan}]{astra}
Giannis Karamanolakis, Subhabrata Mukherjee, Guoqing Zheng, and Ahmed Hassan.
  2021.
\newblock Self-training with weak supervision.
\newblock In \emph{Proceedings of the 2021 Conference of the North American
  Chapter of the Association for Computational Linguistics: Human Language
  Technologies}, pages 845--863.

\bibitem[{Kaushal et~al.(2019{\natexlab{a}})Kaushal, Iyer, Doctor, Sahoo,
  Dubal, Kothawade, Mahadev, Dargan, and
  Ramakrishnan}]{kaushal2019demystifying}
Vishal Kaushal, Rishabh Iyer, Khoshrav Doctor, Anurag Sahoo, Pratik Dubal,
  Suraj Kothawade, Rohan Mahadev, Kunal Dargan, and Ganesh Ramakrishnan.
  2019{\natexlab{a}}.
\newblock Demystifying multi-faceted video summarization: tradeoff between
  diversity, representation, coverage and importance.
\newblock In \emph{2019 IEEE Winter Conference on Applications of Computer
  Vision (WACV)}, pages 452--461. IEEE.

\bibitem[{Kaushal et~al.(2019{\natexlab{b}})Kaushal, Iyer, Kothawade, Mahadev,
  Doctor, and Ramakrishnan}]{kaushal2019learning}
Vishal Kaushal, Rishabh Iyer, Suraj Kothawade, Rohan Mahadev, Khoshrav Doctor,
  and Ganesh Ramakrishnan. 2019{\natexlab{b}}.
\newblock Learning from less data: A unified data subset selection and active
  learning framework for computer vision.
\newblock In \emph{2019 IEEE Winter Conference on Applications of Computer
  Vision (WACV)}, pages 1289--1299. IEEE.

\bibitem[{Kothawade et~al.(2021)Kothawade, Kaushal, Ramakrishnan, Bilmes, and
  Iyer}]{kothawade2021prism}
Suraj Kothawade, Vishal Kaushal, Ganesh Ramakrishnan, Jeff Bilmes, and Rishabh
  Iyer. 2021.
\newblock Prism: A rich class of parameterized submodular information measures
  for guided subset selection.
\newblock \emph{arXiv preprint arXiv:2103.00128}.

\bibitem[{Li and Roth(2002)}]{trec}
Xin Li and Dan Roth. 2002.
\newblock Learning question classifiers.
\newblock In \emph{COLING 2002: The 19th International Conference on
  Computational Linguistics}.

\bibitem[{Maheshwari et~al.(2021)Maheshwari, Chatterjee, Killamsetty, Iyer, and
  Ramakrishnan}]{spear}
Ayush Maheshwari, Oishik Chatterjee, KrishnaTeja Killamsetty, Rishabh~K. Iyer,
  and Ganesh Ramakrishnan. 2021.
\newblock \href {http://arxiv.org/abs/2008.09887} {Data programming using
  semi-supervision and subset selection}.
\newblock In \emph{Proceedings of the 59th Annual Meeting of the Association
  for Computational Linguistics}.

\bibitem[{Maheshwari et~al.(2022)Maheshwari, Killamsetty, Ramakrishnan, Iyer,
  Danilevsky, and Popa}]{maheshwari2022learning}
Ayush Maheshwari, Krishnateja Killamsetty, Ganesh Ramakrishnan, Rishabh Iyer,
  Marina Danilevsky, and Lucian Popa. 2022.
\newblock Learning to robustly aggregate labeling functions for semi-supervised
  data programming.
\newblock In \emph{Findings of the Association for Computational Linguistics:
  ACL 2022}, pages 1188--1202.

\bibitem[{Mintz et~al.(2009)Mintz, Bills, Snow, and
  Jurafsky}]{mintz2009distant}
Mike Mintz, Steven Bills, Rion Snow, and Dan Jurafsky. 2009.
\newblock Distant supervision for relation extraction without labeled data.
\newblock In \emph{Proceedings of the Joint Conference of the 47th Annual
  Meeting of the ACL and the 4th International Joint Conference on Natural
  Language Processing of the AFNLP}, pages 1003--1011.

\bibitem[{Pryzant et~al.(2022)Pryzant, Yang, Xu, Zhu, and
  Zeng}]{pryzant2022automatic}
Reid Pryzant, Ziyi Yang, Yichong Xu, Chenguang Zhu, and Michael Zeng. 2022.
\newblock Automatic rule induction for efficient semi-supervised learning.
\newblock \emph{arXiv preprint arXiv:2205.09067}.

\bibitem[{Ren et~al.(2018)Ren, Zeng, Yang, and Urtasun}]{l2r}
Mengye Ren, Wenyuan Zeng, Bin Yang, and Raquel Urtasun. 2018.
\newblock Learning to reweight examples for robust deep learning.
\newblock In \emph{International Conference on Machine Learning}, pages
  4334--4343.

\bibitem[{Sahay et~al.(2021)Sahay, Nasery, Maheshwari, Ramakrishnan, and
  Iyer}]{sahay2021rule}
Atul Sahay, Anshul Nasery, Ayush Maheshwari, Ganesh Ramakrishnan, and Rishabh~K
  Iyer. 2021.
\newblock Rule augmented unsupervised constituency parsing.
\newblock In \emph{ACL/IJCNLP (Findings)}.

\bibitem[{Sen et~al.(2020)Sen, Danilevsky, Li, Brahma, Boehm, Chiticariu, and
  Krishnamurthy}]{sen2020learning}
Prithviraj Sen, Marina Danilevsky, Yunyao Li, Siddhartha Brahma, Matthias
  Boehm, Laura Chiticariu, and Rajasekar Krishnamurthy. 2020.
\newblock Learning explainable linguistic expressions with neural inductive
  logic programming for sentence classification.
\newblock In \emph{Proceedings of the 2020 Conference on Empirical Methods in
  Natural Language Processing (EMNLP)}, pages 4211--4221.

\bibitem[{Settles(2009)}]{settles2009active}
Burr Settles. 2009.
\newblock Active learning literature survey.

\bibitem[{Shnarch et~al.(2017)Shnarch, Levy, Raykar, and
  Slonim}]{shnarch2017Grasp}
Eyal Shnarch, Ran Levy, Vikas Raykar, and Noam Slonim. 2017.
\newblock Grasp: Rich patterns for argumentation mining.
\newblock In \emph{Proceedings of the 2017 Conference on Empirical Methods in
  Natural Language Processing}, pages 1345--1350.

\bibitem[{Singh et~al.(2023)Singh, Subramanian, Maheshwari, Narayan, Shetty,
  and Ramakrishnan}]{singh2023eigen}
Abhishek Singh, Venkatapathy Subramanian, Ayush Maheshwari, Pradeep Narayan,
  Devi~Prasad Shetty, and Ganesh Ramakrishnan. 2023.
\newblock Eigen: Expert-informed joint learning aggregation for high-fidelity
  information extraction from document images.
\newblock In \emph{Machine Learning for Health (ML4H)}, pages 559--573. PMLR.

\bibitem[{Sivasubramanian et~al.(2023)Sivasubramanian, Maheshwari, Prathosh,
  Shenoy, and Ramakrishnan}]{sivasubramanian2023adaptive}
Durga Sivasubramanian, Ayush Maheshwari, AP~Prathosh, Pradeep Shenoy, and
  Ganesh Ramakrishnan. 2023.
\newblock Adaptive mixing of auxiliary losses in supervised learning.
\newblock In \emph{Proceedings of the AAAI Conference on Artificial
  Intelligence}, volume~37, pages 9855--9863.

\bibitem[{Socher et~al.(2013)Socher, Perelygin, Wu, Chuang, Manning, Ng, and
  Potts}]{sst5}
Richard Socher, Alex Perelygin, Jean Wu, Jason Chuang, Christopher~D Manning,
  Andrew~Y Ng, and Christopher Potts. 2013.
\newblock Recursive deep models for semantic compositionality over a sentiment
  treebank.
\newblock In \emph{Proceedings of the 2013 conference on empirical methods in
  natural language processing}, pages 1631--1642.

\bibitem[{Van~Engelen and Hoos(2020)}]{van2020survey}
Jesper~E Van~Engelen and Holger~H Hoos. 2020.
\newblock A survey on semi-supervised learning.
\newblock \emph{Machine Learning}, 109(2):373--440.

\bibitem[{Varma and R\'{e}(2018)}]{snuba}
Paroma Varma and Christopher R\'{e}. 2018.
\newblock \href {https://doi.org/10.14778/3291264.3291268} {Snuba: Automating
  weak supervision to label training data}.
\newblock \emph{Proc. VLDB Endow.}, 12(3):223–236.

\bibitem[{Varma and R{\'e}(2018)}]{varma2018snuba}
Paroma Varma and Christopher R{\'e}. 2018.
\newblock Snuba: automating weak supervision to label training data.
\newblock In \emph{Proceedings of the VLDB Endowment. International Conference
  on Very Large Data Bases}, volume~12, page 223. NIH Public Access.

\bibitem[{Wei et~al.(2015)Wei, Iyer, and Bilmes}]{wei2015submodularity}
Kai Wei, Rishabh Iyer, and Jeff Bilmes. 2015.
\newblock Submodularity in data subset selection and active learning.
\newblock In \emph{International Conference on Machine Learning}, pages
  1954--1963.

\bibitem[{Wilcoxon(1992)}]{wilcoxon1992individual}
Frank Wilcoxon. 1992.
\newblock Individual comparisons by ranking methods.
\newblock In \emph{Breakthroughs in statistics}, pages 196--202. Springer.

\bibitem[{Yao et~al.(2017)Yao, Wan, and Xiao}]{yao2017recent}
Jin-ge Yao, Xiaojun Wan, and Jianguo Xiao. 2017.
\newblock Recent advances in document summarization.
\newblock \emph{Knowledge and Information Systems}, 53(2):297--336.

\end{thebibliography}
\newpage
\appendix
\newpage
{\huge{Appendix}}

\section{Non-submodularity of $f_{PCA}$}
We observed that $f_{PCA}$ variant of FAIR does not follow the submodularity properties due to which we could not find any approximation guarantees for $f_{PCA}$. We illustrate this using the following example.
\theoremstyle{definition}
\newtheorem{exmp}{Example}[section]
\begin{exmp}\label{ex: 1}
    Consider sets $S$ and $T$ such that $S\subset T\subset \mathcal{R}$. Let us consider $S = \{R_1\}$ and $T = \{R_1, R_2\}$ and consider the rule $R_3\in \mathcal{R} \neq R_1, R_2$ where $\alpha(R_1) = \alpha(R_2) = 1.0$ and $\alpha(R_3) = 0.5$ and all the rules covers same points with $\beta(R_1) = \beta(R_2) = \beta(R_3) = 0.1$ and $\mu(R_1) = 1.0, \mu(R_1\cup R_2) = 0.5, \mu(R_1\cup R_3) = 0.5 \text{ and } \mu(R_1\cup R_2 \cup R_3) = 0.5$ then $f_{PCA}$ violates the property of diminishing marginal returns.
\end{exmp}

We first calculate $f_{PCA}(S) = 1.0 + 0.1 + 1.0 = 2.1$ and $f_{PCA}(S\cup R_3) = (1.0 + 0.5)/2 + 0.1 + 0.5 = 0.90$. Then we calculate $f_{PCA}(T) = (1.0+1.0)/2+0.1+0.5 = 1.6$. $f_{PCA}(T\cup R_3) = (1+1+0.5)/3+0.1+0.5= 1.43$.

Now observe that $f_{PCA}(S\cup R_3 )- f_{PCA}(S) = -1.2 \leq f_{PCA}(T\cup R_3 )- f_{PCA}(T) = -0.2$. Hence $f_{PCA}$ does not follows the property of diminishing returns on this example. Hence we conclude that $f_{PCA}$ is not submodular.

\begin{table}

\begin{tabular}{|c|c|c|c|c|}
\hline
\textbf{Dataset} & \textbf{|$\mathcal{L}$|} & \textbf{|$\mathcal{U}$|} & \textbf{|TEST|}  \\ \hline
\textbf{IMDB}    & 71           & 1278         & 500             \\ \hline
\textbf{YouTube} & 54           & 977          & 500             \\ \hline
\textbf{SMS}     & 463          & 8335         & 500             \\ \hline
\textbf{TREC}    & 273          & 4918         & 500             \\ \hline
\textbf{SST}     & 568          & 10219        & 500             \\ \hline
\end{tabular}
\caption{More details on size of labeled set $\mathcal{L}$, unlabeled set $\mathcal{U}$ and size of Test data.}
\label{tab: dataset}
\end{table}

\begin{table}
\begin{tabular}
{|l|r|r|r|}
\hline
        & \multicolumn{1}{l|}{Snuba} & \multicolumn{1}{l|}{M-Grasp} & \multicolumn{1}{l|}{Classifier} \\ \hline
$p$-value & 0.0018                     & 0.0265                       & 0.0180                          \\ \hline
$z$-value & 2.9000                     & 1.9300                       & 2.0900                          \\ \hline
\end{tabular}
\caption{p-value and z-value for statistical significance tests}
\label{tab:Significance tests}
\end{table}

\begin{table*}[]
\centering
\caption{Final committed set of rules generated using Classifier weights for the YouTube dataset.}

\label{tab: QA1}
\begin{tabular}{|c|cccccc|}
\hline
                       & \multicolumn{6}{c|}{ARI: CW, Dataset: YouTube}                                                                                                                                                                            \\ \hline
Filtering              & \multicolumn{2}{c|}{CW}                                                   & \multicolumn{2}{c|}{GC}                                                 & \multicolumn{2}{c|}{PCA}                          \\ \hline
Class                  & \multicolumn{1}{c|}{Spam}               & \multicolumn{1}{c|}{Ham}              & \multicolumn{1}{c|}{Spam}               & \multicolumn{1}{c|}{Ham}            & \multicolumn{1}{c|}{Spam}               & Ham           \\ \hline
\multirow{6}{*}{Rules} & \multicolumn{1}{c|}{\textit{check}}     & \multicolumn{1}{c|}{\textit{song}}    & \multicolumn{1}{c|}{\textit{check}}     & \multicolumn{1}{c|}{\textit{song}}  & \multicolumn{1}{c|}{\textit{check}}     & \textit{song} \\ \cline{2-7} 
                       & \multicolumn{1}{c|}{\textit{subscribe}} & \multicolumn{1}{c|}{\textit{views}}   & \multicolumn{1}{c|}{\textit{subscribe}} & \multicolumn{1}{c|}{\textit{good}}  & \multicolumn{1}{c|}{\textit{subscribe}} & \textit{\_}   \\ \cline{2-7} 
                       & \multicolumn{1}{c|}{\textit{com}}       & \multicolumn{1}{c|}{\textit{shuffle}} & \multicolumn{1}{c|}{\textit{http}}      & \multicolumn{1}{c|}{\textit{years}} & \multicolumn{1}{c|}{\textit{http}}      & \textit{\_}   \\ \cline{2-7} 
                       & \multicolumn{1}{c|}{\textit{channel}}   & \multicolumn{1}{c|}{\textit{\_}}      & \multicolumn{1}{c|}{\textit{channel}}   & \multicolumn{1}{c|}{\textit{\_}}    & \multicolumn{1}{c|}{\textit{channel}}   & \textit{\_}   \\ \cline{2-7} 
                       & \multicolumn{1}{c|}{\textit{\_}}        & \multicolumn{1}{c|}{\textit{\_}}      & \multicolumn{1}{c|}{\textit{\_}}        & \multicolumn{1}{c|}{\textit{\_}}    & \multicolumn{1}{c|}{\textit{watch}}     & \textit{\_}   \\ \cline{2-7} 
                       & \multicolumn{1}{c|}{\textit{\_}}        & \multicolumn{1}{c|}{\textit{\_}}      & \multicolumn{1}{c|}{\textit{\_}}        & \multicolumn{1}{c|}{\textit{\_}}    & \multicolumn{1}{c|}{\textit{com}}       & \textit{\_}   \\ \hline
\end{tabular}
\label{tab:final-rules}
\end{table*}


\begin{figure}
    \centering
    \includegraphics[scale = 0.40]{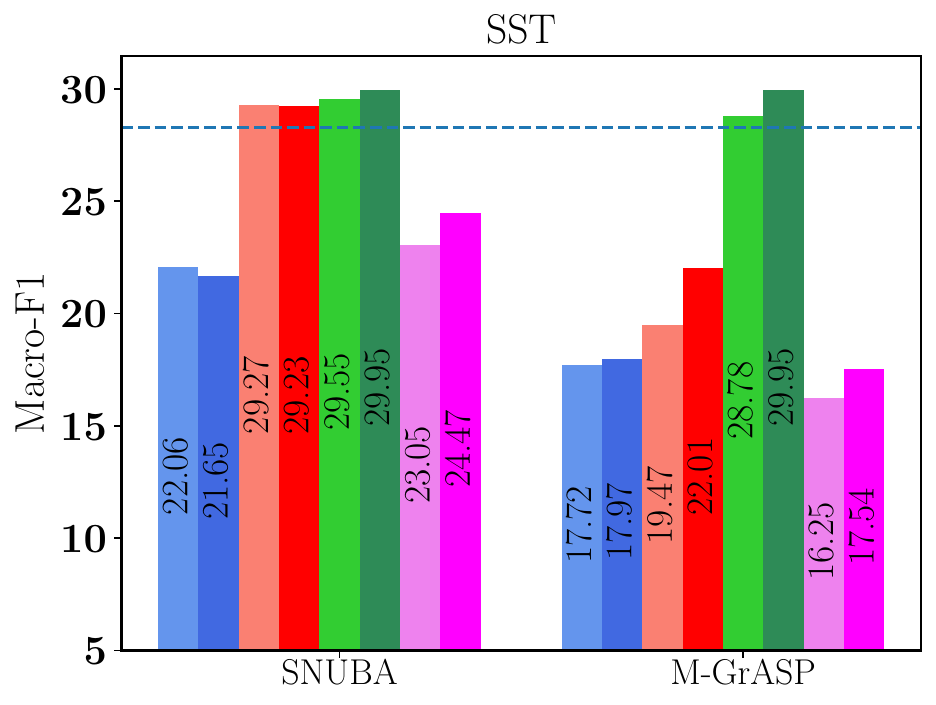}
    \caption{Results of $f_{PCA}$ on SST dataset}
    \label{fig:res10}
\end{figure}

\begin{table*}[]
\centering
\caption{Rule statistics for Snuba and  GC variant of \our{}. \#RulesCS is the number of rules in the candidate set, and \#Rules are the number of rules in the committed set. Coverage is the percentage of points covered in the unlabeled set, Agreement is the percentage of points on which rules have non-conflicting labels and Precision is the micro-precision of rules on the test set. Number of rules in the final set was the same for Snuba and \our{}.}
\begin{tabular}{|l|c|c|cc|cc|cc|}
\hline
\textbf{Dataset} & \multicolumn{1}{c|}{\textbf{\#RulesCS}} & \multicolumn{1}{c|}{\textbf{\#Rules}} & \multicolumn{2}{c|}{\textbf{~~Coverage~~}} & \multicolumn{2}{c|}{\textbf{~~Agreement~~}} & \multicolumn{2}{c|}{\textbf{~~Precision~~}} \\ \hline
                 &                                         &                                       & \multicolumn{1}{l|}{~~Snuba~~}    & ~~GC~~     & \multicolumn{1}{l|}{~~Snuba~~}    & ~~GC~~      & \multicolumn{1}{l|}{~~Snuba~~}    & ~~GC~~      \\ \hline
\textbf{YouTube} & 99                                      & 7                                     & \multicolumn{1}{c|}{55.8}     & 54.3   & \multicolumn{1}{c|}{95.8}     & 96.0      & \multicolumn{1}{c|}{94.3}     & 94.5    \\ \hline
\textbf{IMDB}    & 143                                     & 16                                    & \multicolumn{1}{c|}{41.2}     & 43.6   & \multicolumn{1}{c|}{95.3}     & 90.2    & \multicolumn{1}{c|}{76.5}     & 77.3    \\ \hline
\textbf{Trec}    & 930                                     & 15                                    & \multicolumn{1}{c|}{62.2}     & 71.9   & \multicolumn{1}{c|}{89.7}     & 87.8    & \multicolumn{1}{c|}{70.1}     & 75.8    \\ \hline
\textbf{SMS}     & 137                                     & 18                                    & \multicolumn{1}{c|}{46.6}     & 41.0     & \multicolumn{1}{c|}{96.6}     & 100     & \multicolumn{1}{c|}{93.8}     & 96.3    \\ \hline
\textbf{SST}    & 2057                                    & 70                                    & \multicolumn{1}{c|}{42.5}     & 40.0     & \multicolumn{1}{c|}{86.5}     & 93.4    & \multicolumn{1}{c|}{35.9}     & 36.1    \\ \hline
\end{tabular}

\label{tab:stats-gc}

\end{table*}

\begin{table*}[]
\centering
\caption{Rule statistics for M-Grasp and GC variant of \our{}.}
\begin{tabular}{|l|c|c|cc|cc|cc|}
\hline
\textbf{Dataset} & \multicolumn{1}{c|}{\textbf{\#RulesCS}} & \multicolumn{1}{c|}{\textbf{\#Rules}} & \multicolumn{2}{c|}{\textbf{~~Coverage~~}} & \multicolumn{2}{c|}{\textbf{~~Agreement~~}} & \multicolumn{2}{c|}{\textbf{~~Precision~~}} \\ \hline
                 &                                         &                                       & \multicolumn{1}{c|}{~M-Grasp~}  & ~~GC~~     & \multicolumn{1}{c|}{~M-Grasp~}   & ~~GC~~     & \multicolumn{1}{c|}{~M-Grasp~}   & ~~GC~~     \\ \hline
\textbf{YouTube} & 200                                     & 7                                     & \multicolumn{1}{c|}{60.1}     & 61.9   & \multicolumn{1}{c|}{88.2}      & 85.2   & \multicolumn{1}{c|}{88.9}      & 91.0     \\ \hline
\textbf{IMDB}    & 200                                     & 16                                    & \multicolumn{1}{c|}{49.3}     & 77.9  & \multicolumn{1}{c|}{89.4}      & 70.3   & \multicolumn{1}{c|}{68.5}      & 69.3   \\ \hline
\textbf{Trec}    & 200                                     & 15                                    & \multicolumn{1}{c|}{47.2}     & 48.9   & \multicolumn{1}{c|}{97.7}      & 99.1   & \multicolumn{1}{c|}{45.8}      & 46.3   \\ \hline
\textbf{SMS}     & 200                                     & 18                                    & \multicolumn{1}{c|}{65.5}     & 71.3   & \multicolumn{1}{c|}{95.5}      & 95.2   & \multicolumn{1}{c|}{84.9}      & 85.1   \\ \hline
\textbf{SST}    & 200                                     & 70                                    & \multicolumn{1}{c|}{90.6}     & 92.1   & \multicolumn{1}{c|}{0.01}      & 0.01   & \multicolumn{1}{c|}{31.2}      & 32.4   \\ \hline
\end{tabular}
\label{tab:stats-grasp}
\end{table*}

\begin{table*}[]
\label{tab:table_class}
\caption{Rule statistics  for Classifier Weights (CW) and  GC variant of \our{}.}
\centering
\begin{tabular}{|l|c|c|cc|cc|cc|}
\hline
\textbf{Dataset} & \multicolumn{1}{c|}{\textbf{\#RulesCS}} & \multicolumn{1}{c|}{\textbf{\#Rules}} & \multicolumn{2}{c|}{\textbf{~~Coverage~~}} & \multicolumn{2}{c|}{\textbf{~~Agreement~~}} & \multicolumn{2}{c|}{\textbf{~~Precision~~}} \\ \hline
                 &                                         &                                       & \multicolumn{1}{l|}{~~CW~~}      & GC      & \multicolumn{1}{l|}{~~CW~~}       & GC      & \multicolumn{1}{l|}{~~CW~~}       & GC      \\ \hline
\textbf{YouTube} & 50                                      & 7                                     & \multicolumn{1}{c|}{63.6}    & 63.4    & \multicolumn{1}{c|}{91.0}       & 93.7    & \multicolumn{1}{c|}{93.3}     & 94.2    \\ \hline
\textbf{IMDB}    & 50                                      & 16                                    & \multicolumn{1}{c|}{60.2}    & 54.3    & \multicolumn{1}{c|}{92.6}     & 89.6    & \multicolumn{1}{c|}{71.9}     & 78.4    \\ \hline
\textbf{Trec}    & 50                                      & 15                                    & \multicolumn{1}{c|}{22.1}    & 13.9    & \multicolumn{1}{c|}{93.2}     & 98.8    & \multicolumn{1}{c|}{76.4}     & 81.4    \\ \hline
\textbf{SMS}     & 50                                      & 18                                    & \multicolumn{1}{c|}{65.7}    & 59.6    & \multicolumn{1}{c|}{73.8}     & 69.5    & \multicolumn{1}{c|}{47.8}     & 48.6    \\ \hline
\textbf{SST}    & 100                                     & 70                                    & \multicolumn{1}{c|}{24.4}    & 15.0      & \multicolumn{1}{c|}{89.5}     & 94.1    & \multicolumn{1}{c|}{29.1}     & 29.2    \\ \hline
\end{tabular}
\label{tab:stats-cw}
\end{table*}

\section{Rule-related statistics}
In tables \ref{tab:stats-gc}, \ref{tab:stats-grasp} and \ref{tab:stats-cw}, we provide the rule-related statistics i.e. Precision of the rules on the test set, Coverage of the rule set as well as the Agreement of the rule set. We observe from the tables that if \our{} has a gain in the coverage then it has a low agreement which is intuitive as more coverage will surely cause more conflicts. The committed set of rules is taken as generated by the respective ARIs. For a fair comparison, we have taken the same number of rules in the final committed set of all filtering approaches.

Observe that although there are mixed results for coverage and agreement as if \our{} GC does better in terms of Agreement, it gets a hit in coverage and vice-versa. However, \our{} GC performs consistently better in terms of precision on the test set across all the rule induction approaches.

\section{Qualitative Analysis}\label{rul:Rules}
We qualitatively analyse the rules generated by different ARI approaches for various dataset in tables  \ref{tab: QA2}, \ref{tab: QA3}, \ref{tab: QA4}, \ref{tab: QA5}, \ref{tab: QA6} and \ref{tab: QA7}. We provide rules in the committed set of different rule filtering approaches. We observe in table \ref{tab: QA2} that FAIR GC provides more diverse rules hence covers both the classes equally in this case while for classifier weights rules are more biased towards the \textit{"ham"} class. Observe that rules make more sense when filtered using the FAIR GC. Similar observations are made in table (\ref{tab: QA4}, \ref{tab: QA5}) as well as table (\ref{tab: QA6}, \ref{tab: QA7}).

\begin{table*}[]
\centering
\caption{Final set of rules generated by CW and FAIR-GC for YouTube dataset.}
\label{tab: QA2}
\begin{tabular}{|c|cc|cc|}
\hline
Filtering               & \multicolumn{2}{c|}{CW}                                           & \multicolumn{2}{c|}{GC}                                 \\ \hline
Class                   & \multicolumn{1}{c|}{Spam}               & Ham                     & \multicolumn{1}{c|}{Spam}              & Ham            \\ \hline
\multirow{14}{*}{Rules} & \multicolumn{1}{c|}{\textit{txt}}       & \textit{yo}             & \multicolumn{1}{c|}{\textit{txt}}      & \textit{ll}    \\ \cline{2-5} 
                        & \multicolumn{1}{c|}{\textit{voucher}}   & \textit{wat}            & \multicolumn{1}{c|}{\textit{mobile}}   & \textit{wat}   \\ \cline{2-5} 
                        & \multicolumn{1}{c|}{\textit{150p}}      & \textit{oh}             & \multicolumn{1}{c|}{\textit{claim}}    & \textit{oh}    \\ \cline{2-5} 
                        & \multicolumn{1}{c|}{\textit{nokia}}     & \textit{ll}             & \multicolumn{1}{c|}{\textit{www}}      & \textit{said}  \\ \cline{2-5} 
                        & \multicolumn{1}{c|}{\textit{ringtones}} & \textit{lol}            & \multicolumn{1}{c|}{\textit{service}}  & \textit{later} \\ \cline{2-5} 
                        & \multicolumn{1}{c|}{\textit{500}}       & \textit{yup}            & \multicolumn{1}{c|}{\textit{uk}}       & \textit{town}  \\ \cline{2-5} 
                        & \multicolumn{1}{c|}{\textit{\_}}        & \textit{haha}           & \multicolumn{1}{c|}{\textit{text}}     & \textit{class} \\ \cline{2-5} 
                        & \multicolumn{1}{c|}{\textit{\_}}        & \textit{ve}             & \multicolumn{1}{c|}{\textit{urgent}}   & \textit{didn}  \\ \cline{2-5} 
                        & \multicolumn{1}{c|}{\textit{\_}}        & \textit{aight}          & \multicolumn{1}{c|}{\textit{ringtone}} & \textit{aight} \\ \cline{2-5} 
                        & \multicolumn{1}{c|}{\textit{\_}}        & \textit{said}           & \multicolumn{1}{c|}{\textit{orange}}   & \textit{gonna} \\ \cline{2-5} 
                        & \multicolumn{1}{c|}{\textit{\_}}        & \textit{sure}           & \multicolumn{1}{c|}{\textit{\_}}       & \textit{\_}    \\ \cline{2-5} 
                        & \multicolumn{1}{c|}{\textit{\_}}        & \textit{fine}           & \multicolumn{1}{c|}{\textit{\_}}       & \textit{\_}    \\ \cline{2-5} 
                        & \multicolumn{1}{c|}{\textit{\_}}        & \textit{sir}            & \multicolumn{1}{c|}{\textit{\_}}       & \textit{\_}    \\ \cline{2-5} 
                        & \multicolumn{1}{c|}{\textit{\_}}        & \textit{later}          & \multicolumn{1}{c|}{\textit{\_}}       & \textit{\_}    \\ \hline
\end{tabular}
\end{table*}

\begin{table*}[]
\centering
\caption{Final set of rules generated by Snuba and \our{}-GC for IMDB dataset.}
\label{tab: QA3}

\begin{tabular}{|c|cc|cc|}
\hline
Filtering              & \multicolumn{2}{c|}{Snuba}                                        & \multicolumn{2}{c|}{GC}                                     \\ \hline
Class                  & \multicolumn{1}{c|}{Action}                & Romance                     & \multicolumn{1}{c|}{Action}               & Romance               \\ \hline
\multirow{9}{*}{Rules} & \multicolumn{1}{c|}{\textit{love}}       & \textit{team}           & \multicolumn{1}{c|}{\textit{york}}      & \textit{world}    \\ \cline{2-5} 
                       & \multicolumn{1}{c|}{\textit{man}}        & \textit{government}     & \multicolumn{1}{c|}{\textit{new york}}  & \textit{year}     \\ \cline{2-5} 
                       & \multicolumn{1}{c|}{\textit{boyfriend}}  & \textit{agent}          & \multicolumn{1}{c|}{\textit{girl}}      & \textit{american} \\ \cline{2-5} 
                       & \multicolumn{1}{c|}{\textit{discovers}}  & \textit{race}           & \multicolumn{1}{c|}{\textit{story}}     & \textit{war}      \\ \cline{2-5} 
                       & \multicolumn{1}{c|}{\textit{friend}}     & \textit{home}           & \multicolumn{1}{c|}{\textit{falls}}     & \textit{agent}    \\ \cline{2-5} 
                       & \multicolumn{1}{c|}{\textit{town}}       & \textit{cop}            & \multicolumn{1}{c|}{\textit{friend}}    & \textit{time}     \\ \cline{2-5} 
                       & \multicolumn{1}{c|}{\textit{friendship}} & \textit{earth}          & \multicolumn{1}{c|}{\textit{best}}      & \textit{team}     \\ \cline{2-5} 
                       & \multicolumn{1}{c|}{\textit{story}}      & \textit{\_}             & \multicolumn{1}{c|}{\textit{meets}}     & \textit{\_}       \\ \cline{2-5} 
                       & \multicolumn{1}{c|}{\textit{falls}}      & \textit{\_}             & \multicolumn{1}{c|}{\textit{young man}} & \textit{\_}       \\ \hline
\end{tabular}
\end{table*}

\begin{table*}[]
\centering
\caption{Final set of rules generated by Classifier Weights (CW) approach on SST Dataset}
\label{tab: QA4}
\begin{tabular}{|c|ccccc|}
\hline
Filtering               & \multicolumn{5}{c|}{CW}                                                                                                                                                                                  \\ \hline
Class                   & \multicolumn{1}{c|}{Negative}         & \multicolumn{1}{c|}{Somewhat negative}  & \multicolumn{1}{c|}{Neutral}               & \multicolumn{1}{c|}{Somewhat positive}            & Positive              \\ \hline
\multirow{22}{*}{Rules} & \multicolumn{1}{c|}{\textit{money}}   & \multicolumn{1}{c|}{\textit{heavy}}     & \multicolumn{1}{c|}{\textit{appealing}}    & \multicolumn{1}{c|}{\textit{era}}                 & \textit{excellent}    \\ \cline{2-6} 
                        & \multicolumn{1}{c|}{\textit{bad}}     & \multicolumn{1}{c|}{\textit{popcorn}}   & \multicolumn{1}{c|}{\textit{conciousness}} & \multicolumn{1}{c|}{\textit{reality}}             & \textit{emotionally}  \\ \cline{2-6} 
                        & \multicolumn{1}{c|}{\textit{assed}}   & \multicolumn{1}{c|}{\textit{confusing}} & \multicolumn{1}{c|}{\textit{glib}}         & \multicolumn{1}{c|}{\textit{presents}}            & \textit{mesmerizing}  \\ \cline{2-6} 
                        & \multicolumn{1}{c|}{\textit{dull}}    & \multicolumn{1}{c|}{\textit{sappy}}     & \multicolumn{1}{c|}{\textit{insomnia}}     & \multicolumn{1}{c|}{\textit{urban}}               & \textit{stuck}        \\ \cline{2-6} 
                        & \multicolumn{1}{c|}{\textit{flaccid}} & \multicolumn{1}{c|}{\textit{feel good}} & \multicolumn{1}{c|}{\textit{movie work}}   & \multicolumn{1}{c|}{\textit{issues}}              & \textit{roles}        \\ \cline{2-6} 
                        & \multicolumn{1}{c|}{\textit{town}}    & \multicolumn{1}{c|}{\textit{\_}}        & \multicolumn{1}{c|}{\textit{just like}}    & \multicolumn{1}{c|}{\textit{riveting}}            & \textit{frailty}      \\ \cline{2-6} 
                        & \multicolumn{1}{c|}{\textit{\_}}      & \multicolumn{1}{c|}{\textit{\_}}        & \multicolumn{1}{c|}{\textit{film ca}}      & \multicolumn{1}{c|}{\textit{liked}}               & \textit{performances} \\ \cline{2-6} 
                        & \multicolumn{1}{c|}{\textit{\_}}      & \multicolumn{1}{c|}{\textit{\_}}        & \multicolumn{1}{c|}{\textit{igby}}         & \multicolumn{1}{c|}{\textit{reality}}             & \textit{\_}           \\ \cline{2-6} 
                        & \multicolumn{1}{c|}{\textit{\_}}      & \multicolumn{1}{c|}{\textit{\_}}        & \multicolumn{1}{c|}{\textit{\_}}           & \multicolumn{1}{c|}{\textit{heart}}               & \textit{\_}           \\ \cline{2-6} 
                        & \multicolumn{1}{c|}{\textit{\_}}      & \multicolumn{1}{c|}{\textit{\_}}        & \multicolumn{1}{c|}{\textit{\_}}           & \multicolumn{1}{c|}{\textit{moved}}               & \textit{\_}           \\ \cline{2-6} 
                        & \multicolumn{1}{c|}{\textit{\_}}      & \multicolumn{1}{c|}{\textit{\_}}        & \multicolumn{1}{c|}{\textit{\_}}           & \multicolumn{1}{c|}{\textit{life}}                & \textit{\_}           \\ \cline{2-6} 
                        & \multicolumn{1}{c|}{\textit{\_}}      & \multicolumn{1}{c|}{\textit{\_}}        & \multicolumn{1}{c|}{\textit{\_}}           & \multicolumn{1}{c|}{\textit{tasty}}               & \textit{\_}           \\ \cline{2-6} 
                        & \multicolumn{1}{c|}{\textit{\_}}      & \multicolumn{1}{c|}{\textit{\_}}        & \multicolumn{1}{c|}{\textit{\_}}           & \multicolumn{1}{c|}{\textit{look}}                & \textit{\_}           \\ \cline{2-6} 
                        & \multicolumn{1}{c|}{\textit{\_}}      & \multicolumn{1}{c|}{\textit{\_}}        & \multicolumn{1}{c|}{\textit{\_}}           & \multicolumn{1}{c|}{\textit{odds}}                & \textit{\_}           \\ \cline{2-6} 
                        & \multicolumn{1}{c|}{\textit{\_}}      & \multicolumn{1}{c|}{\textit{\_}}        & \multicolumn{1}{c|}{\textit{\_}}           & \multicolumn{1}{c|}{\textit{actor}}               & \textit{\_}           \\ \cline{2-6} 
                        & \multicolumn{1}{c|}{\textit{\_}}      & \multicolumn{1}{c|}{\textit{\_}}        & \multicolumn{1}{c|}{\textit{\_}}           & \multicolumn{1}{c|}{\textit{spice}}               & \textit{\_}           \\ \cline{2-6} 
                        & \multicolumn{1}{c|}{\textit{\_}}      & \multicolumn{1}{c|}{\textit{\_}}        & \multicolumn{1}{c|}{\textit{\_}}           & \multicolumn{1}{c|}{\textit{new}}                 & \textit{\_}           \\ \cline{2-6} 
                        & \multicolumn{1}{c|}{\textit{\_}}      & \multicolumn{1}{c|}{\textit{\_}}        & \multicolumn{1}{c|}{\textit{\_}}           & \multicolumn{1}{c|}{\textit{diversion}}           & \textit{\_}           \\ \cline{2-6} 
                        & \multicolumn{1}{c|}{\textit{\_}}      & \multicolumn{1}{c|}{\textit{\_}}        & \multicolumn{1}{c|}{\textit{\_}}           & \multicolumn{1}{c|}{\textit{voices}}              & \textit{\_}           \\ \cline{2-6} 
                        & \multicolumn{1}{c|}{\textit{\_}}      & \multicolumn{1}{c|}{\textit{\_}}        & \multicolumn{1}{c|}{\textit{\_}}           & \multicolumn{1}{c|}{\textit{strong performances}} & \textit{\_}           \\ \cline{2-6} 
                        & \multicolumn{1}{c|}{\textit{\_}}      & \multicolumn{1}{c|}{\textit{\_}}        & \multicolumn{1}{c|}{\textit{\_}}           & \multicolumn{1}{c|}{\textit{answers}}             & \textit{\_}           \\ \cline{2-6} 
                        & \multicolumn{1}{c|}{\textit{\_}}      & \multicolumn{1}{c|}{\textit{\_}}        & \multicolumn{1}{c|}{\textit{\_}}           & \multicolumn{1}{c|}{\textit{works}}               & \textit{\_}           \\ \hline
\end{tabular}
\end{table*}
\begin{table*}[]
\centering
\caption{Final set of rules generated by \our{}-GC for SST dataset}
\label{tab: QA5}
\begin{tabular}{|c|ccccc|}
\hline
Filtering               & \multicolumn{5}{c|}{GC}                                                                                                                                                                                 \\ \hline
Class                   & \multicolumn{1}{c|}{Negative}             & \multicolumn{1}{c|}{Somewhat negative}     & \multicolumn{1}{c|}{Neutral}              & \multicolumn{1}{c|}{Somewhat positive}  & Positive                 \\ \hline
\multirow{18}{*}{Rules} & \multicolumn{1}{c|}{\textit{assed}}       & \multicolumn{1}{c|}{\textit{heavy}}        & \multicolumn{1}{c|}{\textit{appealing}}   & \multicolumn{1}{c|}{\textit{presents}}  & \textit{}                \\ \cline{2-6} 
                        & \multicolumn{1}{c|}{\textit{really bad}}  & \multicolumn{1}{c|}{\textit{popcorn}}      & \multicolumn{1}{c|}{\textit{just like}}   & \multicolumn{1}{c|}{\textit{issues}}    & \textit{mesmerizing}     \\ \cline{2-6} 
                        & \multicolumn{1}{c|}{\textit{assed film}}  & \multicolumn{1}{c|}{\textit{popcorn film}} & \multicolumn{1}{c|}{\textit{movie work}}  & \multicolumn{1}{c|}{\textit{tasty}}     & \textit{stuck}           \\ \cline{2-6} 
                        & \multicolumn{1}{c|}{\textit{nonexistent}} & \multicolumn{1}{c|}{\textit{numbers}}      & \multicolumn{1}{c|}{\textit{work better}} & \multicolumn{1}{c|}{\textit{spice}}     & \textit{graet films}     \\ \cline{2-6} 
                        & \multicolumn{1}{c|}{\textit{week}}        & \multicolumn{1}{c|}{\textit{away}}         & \multicolumn{1}{c|}{\textit{like igby}}   & \multicolumn{1}{c|}{\textit{diversion}} & \textit{tremedous piece} \\ \cline{2-6} 
                        & \multicolumn{1}{c|}{\textit{dull}}        & \multicolumn{1}{c|}{\textit{felt}}         & \multicolumn{1}{c|}{\textit{igby}}        & \multicolumn{1}{c|}{\textit{odds}}      & \textit{leads}           \\ \cline{2-6} 
                        & \multicolumn{1}{c|}{\textit{hours}}       & \multicolumn{1}{c|}{\textit{\_}}           & \multicolumn{1}{c|}{\textit{happy}}       & \multicolumn{1}{c|}{\textit{moved}}     & \textit{thoughtful}      \\ \cline{2-6} 
                        & \multicolumn{1}{c|}{\textit{bad}}         & \multicolumn{1}{c|}{\textit{\_}}           & \multicolumn{1}{c|}{\textit{doing}}       & \multicolumn{1}{c|}{\textit{answers}}   & \textit{\_}              \\ \cline{2-6} 
                        & \multicolumn{1}{c|}{\textit{\_}}          & \multicolumn{1}{c|}{\textit{\_}}           & \multicolumn{1}{c|}{\textit{flawed}}      & \multicolumn{1}{c|}{\textit{simone}}    & \textit{\_}              \\ \cline{2-6} 
                        & \multicolumn{1}{c|}{\textit{\_}}          & \multicolumn{1}{c|}{\textit{\_}}           & \multicolumn{1}{c|}{\textit{\_}}          & \multicolumn{1}{c|}{\textit{perfectly}} & \textit{\_}              \\ \cline{2-6} 
                        & \multicolumn{1}{c|}{\textit{\_}}          & \multicolumn{1}{c|}{\textit{\_}}           & \multicolumn{1}{c|}{\textit{\_}}          & \multicolumn{1}{c|}{\textit{enjoyable}} & \textit{\_}              \\ \cline{2-6} 
                        & \multicolumn{1}{c|}{\textit{\_}}          & \multicolumn{1}{c|}{\textit{\_}}           & \multicolumn{1}{c|}{\textit{\_}}          & \multicolumn{1}{c|}{\textit{quirky}}    & \textit{\_}              \\ \cline{2-6} 
                        & \multicolumn{1}{c|}{\textit{\_}}          & \multicolumn{1}{c|}{\textit{\_}}           & \multicolumn{1}{c|}{\textit{\_}}          & \multicolumn{1}{c|}{\textit{step}}      & \textit{\_}              \\ \cline{2-6} 
                        & \multicolumn{1}{c|}{\textit{\_}}          & \multicolumn{1}{c|}{\textit{\_}}           & \multicolumn{1}{c|}{\textit{\_}}          & \multicolumn{1}{c|}{\textit{heart}}     & \textit{\_}              \\ \cline{2-6} 
                        & \multicolumn{1}{c|}{\textit{\_}}          & \multicolumn{1}{c|}{\textit{\_}}           & \multicolumn{1}{c|}{\textit{\_}}          & \multicolumn{1}{c|}{\textit{french}}    & \textit{\_}              \\ \cline{2-6} 
                        & \multicolumn{1}{c|}{\textit{\_}}          & \multicolumn{1}{c|}{\textit{\_}}           & \multicolumn{1}{c|}{\textit{\_}}          & \multicolumn{1}{c|}{\textit{works}}     & \textit{\_}              \\ \cline{2-6} 
                        & \multicolumn{1}{c|}{\textit{\_}}          & \multicolumn{1}{c|}{\textit{\_}}           & \multicolumn{1}{c|}{\textit{\_}}          & \multicolumn{1}{c|}{\textit{actor}}     & \textit{\_}              \\ \cline{2-6} 
                        & \multicolumn{1}{c|}{\textit{\_}}          & \multicolumn{1}{c|}{\textit{\_}}           & \multicolumn{1}{c|}{\textit{\_}}          & \multicolumn{1}{c|}{\textit{leave}}     & \textit{\_}              \\ \hline
\end{tabular}
\end{table*}
\begin{table*}[]
\centering
\caption{Final set of rules generated by Snuba for SST dataset. We display few selected rules out of total 70 rules.}
\label{tab: QA6}

\begin{tabular}{|c|ccccc|}
                        \hline
Filtering               & \multicolumn{5}{c|}{GC}                                                                                                                                                                           \\ \hline
Class                   & \multicolumn{1}{c|}{Negative}             & \multicolumn{1}{c|}{Somewhat negative} & \multicolumn{1}{c|}{Neutral}              & \multicolumn{1}{c|}{Somewhat positive}      & Positive           \\ \hline
\multirow{10}{*}{Rules} & \multicolumn{1}{c|}{\textit{worse}}       & \multicolumn{1}{c|}{\textit{like it}}  & \multicolumn{1}{c|}{\textit{always}}      & \multicolumn{1}{c|}{\textit{beautiful}}     & \textit{theme}     \\ \cline{2-6} 
                        & \multicolumn{1}{c|}{\textit{week}}        & \multicolumn{1}{c|}{\textit{getting}}  & \multicolumn{1}{c|}{\textit{at time}}     & \multicolumn{1}{c|}{\textit{occasionaly}}   & \textit{epic}      \\ \cline{2-6} 
                        & \multicolumn{1}{c|}{\textit{just}}        & \multicolumn{1}{c|}{\textit{no}}       & \multicolumn{1}{c|}{\textit{appealing}}   & \multicolumn{1}{c|}{\textit{tale of}}       & \textit{not be}    \\ \cline{2-6} 
                        & \multicolumn{1}{c|}{\textit{imagine}}     & \multicolumn{1}{c|}{\textit{by it}}    & \multicolumn{1}{c|}{\textit{work better}} & \multicolumn{1}{c|}{\textit{psychological}} & \textit{excellent} \\ \cline{2-6} 
                        & \multicolumn{1}{c|}{\textit{should have}} & \multicolumn{1}{c|}{\textit{\_}}       & \multicolumn{1}{c|}{\textit{\_}}          & \multicolumn{1}{c|}{\textit{works}}         & \textit{music}     \\ \cline{2-6} 
                        & \multicolumn{1}{c|}{\textit{\_}}          & \multicolumn{1}{c|}{\textit{\_}}       & \multicolumn{1}{c|}{\textit{\_}}          & \multicolumn{1}{c|}{\textit{rock}}          & \textit{\_}        \\ \cline{2-6} 
                        & \multicolumn{1}{c|}{\textit{\_}}          & \multicolumn{1}{c|}{\textit{\_}}       & \multicolumn{1}{c|}{\textit{\_}}          & \multicolumn{1}{c|}{\textit{ernest}}        & \textit{\_}        \\ \cline{2-6} 
                        & \multicolumn{1}{c|}{\textit{\_}}          & \multicolumn{1}{c|}{\textit{\_}}       & \multicolumn{1}{c|}{\textit{\_}}          & \multicolumn{1}{c|}{\textit{the actor}}     & \textit{\_}        \\ \cline{2-6} 
                        & \multicolumn{1}{c|}{\textit{\_}}          & \multicolumn{1}{c|}{\textit{\_}}       & \multicolumn{1}{c|}{\textit{\_}}          & \multicolumn{1}{c|}{\textit{story that}}    & \textit{\_}        \\ \cline{2-6} 
                        & \multicolumn{1}{c|}{\textit{\_}}          & \multicolumn{1}{c|}{\textit{\_}}       & \multicolumn{1}{c|}{\textit{\_}}          & \multicolumn{1}{c|}{\textit{during}}        & \textit{\_}        \\ \hline
\end{tabular}
\end{table*}

\begin{table*}[]
\centering
\caption{Final set of rules generated by \our{}-GC for SST dataset over Snuba generated candidate rule set. We display few selected rules out of total 70 rules.}
\label{tab: QA7}
\begin{tabular}{|c|ccccc|}
                        \hline
Filtering               & \multicolumn{5}{c|}{Snuba}                                                                                                                                                             \\ \hline
Class                   & \multicolumn{1}{c|}{Negative}      & \multicolumn{1}{c|}{Somewhat negative}       & \multicolumn{1}{c|}{Neutral}     & \multicolumn{1}{c|}{Somewhat positive}   & Positive             \\ \hline
\multirow{18}{*}{Rules} & \multicolumn{1}{c|}{\textit{bad}}  & \multicolumn{1}{c|}{\textit{too}}            & \multicolumn{1}{c|}{\textit{see the}} & \multicolumn{1}{c|}{\textit{who}}        & \textit{comedy with} \\ \cline{2-6} 
                        & \multicolumn{1}{c|}{\textit{week}} & \multicolumn{1}{c|}{\textit{even}}           & \multicolumn{1}{c|}{\textit{slow}} & \multicolumn{1}{c|}{\textit{compelling}} & \textit{music}       \\ \cline{2-6} 
                        & \multicolumn{1}{c|}{\textit{just}} & \multicolumn{1}{c|}{\textit{where}}          & \multicolumn{1}{c|}{\textit{\_}} & \multicolumn{1}{c|}{\textit{both}}       & \textit{\_}          \\ \cline{2-6} 
                        & \multicolumn{1}{c|}{\textit{\_}}   & \multicolumn{1}{c|}{\textit{only}}           & \multicolumn{1}{c|}{\textit{\_}} & \multicolumn{1}{c|}{\textit{.}}          & \textit{\_}          \\ \cline{2-6} 
                        & \multicolumn{1}{c|}{\textit{\_}}   & \multicolumn{1}{c|}{\textit{no}}             & \multicolumn{1}{c|}{\textit{\_}} & \multicolumn{1}{c|}{\textit{era}}        & \textit{\_}          \\ \cline{2-6} 
                        & \multicolumn{1}{c|}{\textit{\_}}   & \multicolumn{1}{c|}{\textit{better}}         & \multicolumn{1}{c|}{\textit{\_}} & \multicolumn{1}{c|}{\textit{there are}}  & \textit{\_}          \\ \cline{2-6} 
                        & \multicolumn{1}{c|}{\textit{\_}}   & \multicolumn{1}{c|}{\textit{out}}            & \multicolumn{1}{c|}{\textit{\_}} & \multicolumn{1}{c|}{\textit{his own}}    & \textit{\_}          \\ \cline{2-6} 
                        & \multicolumn{1}{c|}{\textit{\_}}   & \multicolumn{1}{c|}{\textit{book}}           & \multicolumn{1}{c|}{\textit{\_}} & \multicolumn{1}{c|}{\textit{\_}}         & \textit{\_}          \\ \cline{2-6} 
                        & \multicolumn{1}{c|}{\textit{\_}}   & \multicolumn{1}{c|}{\textit{it}}             & \multicolumn{1}{c|}{\textit{\_}} & \multicolumn{1}{c|}{\textit{\_}}         & \textit{\_}          \\ \cline{2-6} 
                        & \multicolumn{1}{c|}{\textit{\_}}   & \multicolumn{1}{c|}{\textit{away}}           & \multicolumn{1}{c|}{\textit{\_}} & \multicolumn{1}{c|}{\textit{\_}}         & \textit{\_}          \\ \cline{2-6} 
                        & \multicolumn{1}{c|}{\textit{\_}}   & \multicolumn{1}{c|}{\textit{but ultimately}} & \multicolumn{1}{c|}{\textit{\_}} & \multicolumn{1}{c|}{\textit{\_}}         & \textit{\_}          \\ \cline{2-6} 
                        & \multicolumn{1}{c|}{\textit{\_}}   & \multicolumn{1}{c|}{\textit{movie to}}       & \multicolumn{1}{c|}{\textit{\_}} & \multicolumn{1}{c|}{\textit{\_}}         & \textit{\_}          \\ \cline{2-6} 
                        & \multicolumn{1}{c|}{\textit{\_}}   & \multicolumn{1}{c|}{\textit{the same}}       & \multicolumn{1}{c|}{\textit{\_}} & \multicolumn{1}{c|}{\textit{\_}}         & \textit{\_}          \\ \cline{2-6} 
                        & \multicolumn{1}{c|}{\textit{\_}}   & \multicolumn{1}{c|}{\textit{sit}}            & \multicolumn{1}{c|}{\textit{\_}} & \multicolumn{1}{c|}{\textit{\_}}         & \textit{\_}          \\ \cline{2-6} 
                        & \multicolumn{1}{c|}{\textit{\_}}   & \multicolumn{1}{c|}{\textit{of an}}          & \multicolumn{1}{c|}{\textit{\_}} & \multicolumn{1}{c|}{\textit{\_}}         & \textit{\_}          \\ \cline{2-6} 
                        & \multicolumn{1}{c|}{\textit{\_}}   & \multicolumn{1}{c|}{\textit{all it}}         & \multicolumn{1}{c|}{\textit{\_}} & \multicolumn{1}{c|}{\textit{\_}}         & \textit{\_}          \\ \cline{2-6} 
                        & \multicolumn{1}{c|}{\textit{\_}}   & \multicolumn{1}{c|}{\textit{try}}            & \multicolumn{1}{c|}{\textit{\_}} & \multicolumn{1}{c|}{\textit{\_}}         & \textit{\_}          \\ \cline{2-6} 
                        & \multicolumn{1}{c|}{\textit{\_}}   & \multicolumn{1}{c|}{\textit{next}}           & \multicolumn{1}{c|}{\textit{\_}} & \multicolumn{1}{c|}{\textit{\_}}         & \textit{\_}          \\ \hline
\end{tabular}
\end{table*}
\end{document}